\pdfoutput=1
\documentclass[10pt,twocolumn]{article}

\usepackage[top=2.5cm,bottom=2.5cm,left=2cm,right=2cm]{geometry}
\usepackage[utf8]{inputenc}
\usepackage[T1]{fontenc}
\usepackage{lmodern}
\usepackage[expansion=false]{microtype}
\usepackage{booktabs}
\usepackage{colortbl}
\usepackage{graphicx}
\usepackage{amsmath,amssymb}
\usepackage{xcolor}
\usepackage{caption}
\usepackage{titlesec}
\usepackage{url}
\usepackage{needspace}
\usepackage{changepage}   %
\usepackage{balance}      %
\usepackage[numbers,sort&compress]{natbib}
\usepackage[breaklinks,colorlinks=true,urlcolor=linkblue,
            citecolor=black,linkcolor=black]{hyperref}

\graphicspath{{figures/}}
\newcommand{\repo}[1]{\path{#1}}

\definecolor{linkblue}{HTML}{2E6DAD}
\definecolor{rulegrey}{HTML}{555555}
\definecolor{rowline}{HTML}{CFCFCF}
\newcommand{\rowsep}{\arrayrulecolor{rowline}\specialrule{0.3pt}{1.4pt}{1.6pt}\arrayrulecolor{black}}

\newcommand{\headingset}{\raggedright\hyphenpenalty=10000\exhyphenpenalty=10000}
\titleformat{\section}{\normalfont\large\bfseries\headingset}{\thesection.}{0.6em}{}
\titleformat{\subsection}{\normalfont\normalsize\bfseries\headingset}{\thesubsection.}{0.6em}{}
\titlespacing*{\section}{0pt}{1.4ex plus .3ex}{0.7ex}
\newcommand{\keyfinding}[1]{%
  \par\needspace{3\baselineskip}\vspace{5pt}%
  {\color{rulegrey}\hrule height 0.4pt}\vspace{4pt}%
  \noindent #1\par\vspace{4pt}%
  {\color{rulegrey}\hrule height 0.4pt}\vspace{6pt}}

\begin{document}

\twocolumn[{%
\begin{center}
  \raisebox{-0.15\height}{\includegraphics[height=9mm]{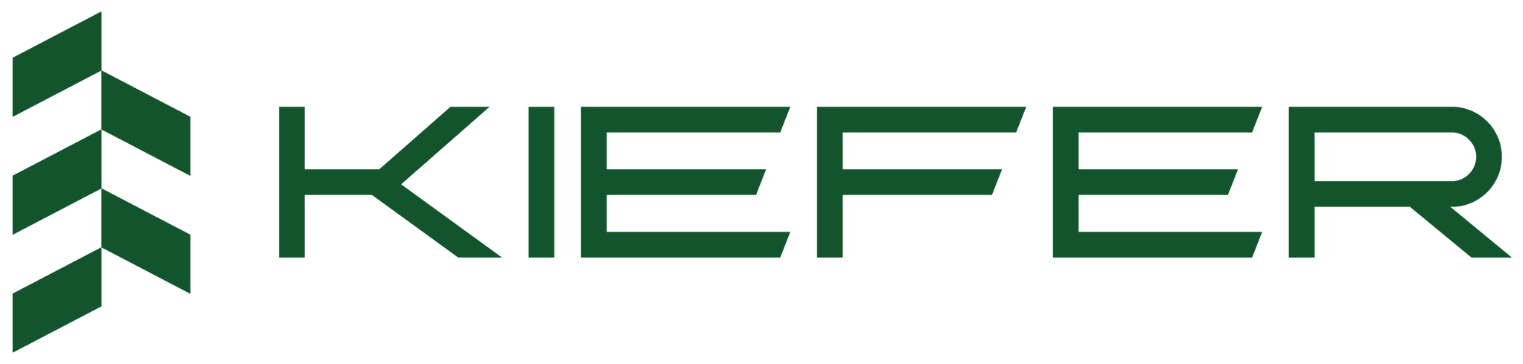}}%
  \hspace{15mm}%
  \raisebox{-0.15\height}{\includegraphics[height=10mm]{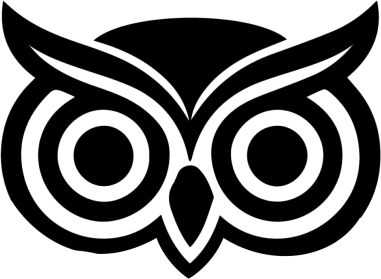}}\\[3.5mm]
  {\fontsize{16}{19}\selectfont\bfseries Thinking in a Low-Resource Language:\\
   What SFT Builds, What RL Fixes, What Accuracy Cannot See}\\[3mm]
  {\fontsize{11}{13}\selectfont\bfseries Ayoub Kirouane\textsuperscript{1}
   \hspace{2em} Christos Petrocheilos\textsuperscript{1}}\\[1.5mm]
  {\fontsize{9.5}{11.5}\selectfont
   \textsuperscript{1}Sophea AI, KIEFER SA, Athens, Greece\\
   \{a.kirouane, c.petrocheilos\}@kiefer.gr\\
   models@sophea.ai}\\[1mm]
  {\fontsize{9}{11}\selectfont\color{rulegrey} Models \& Benchmarks:
   \url{https://huggingface.co/KIEFERSA} \hspace{1em} August 2026}
\end{center}
\vspace{3mm}
\begin{minipage}{\textwidth}
\begin{adjustwidth}{3em}{3em}
\begin{center}{\fontsize{10.5}{12.5}\selectfont\bfseries Abstract}\end{center}
\vspace{1mm}
\small\noindent
Take three frontier mixture-of-experts models (Alibaba, OpenAI, NVIDIA; $3.6$--$4.0$B active
parameters each) and fine-tune them to reason in a low-resource language. On accuracy
benchmarks, almost nothing happens, and the benchmark itself is noise at this scale: changing
only the random seed moves the score by 7.7 points, more than every data and recipe effect we
measured. That null is our first result. The real changes live where accuracy cannot see. Base
models never think in Greek: 0 of 1{,}000 reasoning traces, even when the question is Greek,
so the model answers correctly while reasoning in a form its user cannot read, audit, or
correct. After supervised fine-tuning (SFT), every released checkpoint reasons in the language
of the question on $\sim$98\% of items, one family at $3\times$ fewer tokens, with judged
grammaticality improving on all four models and general ability within a few points of each
model's own base in both languages: nothing was forgotten, and fluency was gained. We propose
the six behavioural dimensions that make such changes measurable (which language the model
reasons in, what it spends, whether it can tell easy from hard, and what it forgot), each gated
to reject any metric that correlates with output length, and we report how our own instruments
lied: six failures, each caught by a control, each of which had already convinced us of
something false. What SFT cannot do is fix its own defects: a quarter of answers skip the
requested format, answers leak into the reasoning channel, and an explicit ``think in English''
is obeyed under half the time. Reinforcement learning with verifiable rewards, pre-registered
before training, fixes the first two outright (answer-format fallback $24\%\!\to\!2.5\%$,
answer-channel leak $3.5\%\!\to\!0.0\%$, both against a flat random-reward control) and
moves the third ($+9.1$\,pp, real but short of its pre-registered bar), while the Greek
reasoning habit survives an accuracy-only gradient untouched ($98.2\%$ fidelity). We release
five checkpoints. The instruments, the controls and the pre-registration travel to any
low-resource language; Greek is the case that let us measure them.
\end{adjustwidth}
\vspace{4mm}
\end{minipage}
}]

\section{Introduction: one number, six dimensions}\label{sec:intro}

A reasoning fine-tune is judged the way every fine-tune is judged: one accuracy number, before and
after. For a low-resource language that number answers a question nobody asked. It does not say
which language the model reasons \emph{in}, how many tokens it burns getting there, whether it can
tell an easy question from a hard one, or what it gave up in exchange.

The study is designed around the constraint a low-resource deployment actually faces: the serving
bill. We hold the \emph{active} parameter count fixed ($3.6$--$4.0$B per token) and vary the lab:
three sparse mixture-of-experts families from Alibaba, OpenAI, and NVIDIA, each a frontier lab's
different architectural bet on the same inference budget (\S\ref{sec:setup}). Sparse MoE is the
architecture of interest precisely because it is cheap: a $20$--$36$B-parameter model that serves
at the cost of a $4$B dense one is what makes local reasoning models economically viable for a
language community that cannot fund frontier-scale inference. Running the same recipe, corpus, and
instruments across all three families is what lets us separate what the fine-tune does from what
one architecture happens to do.

Greek already has a dedicated open-model ecosystem we build on rather than from scratch: the
Meltemi and Krikri instruction models~\citep{voukoutis2024meltemi,roussis2025krikri}, their translated
evaluation suites, and our own Sophea-Titan-1 general-purpose model. What none of that line
measures is long-form \emph{reasoning} behaviour: those models answer in Greek, but no released
checkpoint thinks out loud in it, and no Greek benchmark can currently tell the difference between
a model that does and one that plans in English and translates at the end.

We measured all of those, and the contrast is the paper. On accuracy, almost nothing happened (the
best of the $15$ arms of \S\ref{sec:side} scores $76.5$ against the base's $77.2$, both on the $1{,}000$-item three-axis
Greek probe in think mode), and for several weeks we treated that as a
corpus problem and built five more corpora to fix it.

None of them fixed it, and eventually we ran the experiment that should have come first: we retrained
one configuration, changing nothing but the random seed. The score moved $7.7$ points. Every
difference we had been interpreting (better data selection, a better corpus, a changed training
schedule) was smaller than that.

This paper is what remains after that discovery, and what replaces it, told in the order the
reasoning has to run. Section~\ref{sec:setup} fixes the models, the corpus, and the evaluation
lanes; Section~\ref{sec:metrics} defines the six dimensions before any of them is used.
Section~\ref{sec:variance} establishes the noise floor. Section~\ref{sec:quality} shows that the
properties which \emph{did}
change are not accuracy and are not seed-sensitive: the language the model reasons in, the tokens it
spends, and whether it can tell an easy question from a hard one. Section~\ref{sec:forget} asks
what the fine-tune forgot. One training-recipe comparison mattered enough to keep as a
self-contained side experiment: it explains our answer-format failure mode and is the one
accuracy effect that survives, replicated across $15$ independently trained arms rather than read
off a single run; it is reported on its own terms in
\S\ref{sec:side}. Section~\ref{sec:lied} reports six instrument failures: five controls each of which killed or corrected a finding
we believed, and Section~\ref{sec:gate} states the acceptance rule they taught us.
Section~\ref{sec:langmatch} then repairs the one defect the recipe installs (the language lock)
and Section~\ref{sec:plan} closes the two remaining pre-registered questions. What follows reads
the same evidence per domain (\S\ref{sec:domain}) and per family (\S\ref{sec:families}), and
Sections~\ref{sec:recommend}--\ref{sec:takeaway} state what we can and cannot recommend on it.

\keyfinding{\textbf{The claim.} We find that for a low-resource-language reasoning fine-tune,
accuracy on a translated benchmark is close to uninformative: it is dominated by training noise,
distorted by answer-format effects, and inflated by contamination that standard checks miss. The
behavioural dimensions are stable, large, and measure what the fine-tune was actually for.}

\section{Setup}\label{sec:setup}

\textbf{Models.} Three sparse mixture-of-experts model
families~\citep{shazeer2017moe,fedus2022switch} from three different labs (four checkpoints in
all, since the Nemotron line contributes two generations), chosen because they occupy
the \emph{same serving budget} while making different architectural bets with it:

\begin{table*}[t]\centering\footnotesize
\begin{tabular}{@{}llrrll@{}}
\toprule
lab & base model & total & active & routing & fine-tuned release \\
\midrule
Alibaba & Qwen3.6-35B-A3B & 36.0B & 3.97B & 8 of 256 & Sophea-Qwen3.6-v1 \\
OpenAI & Gpt-OSS-20B & 20.9B & 3.60B & 4 of 32 & Sophea-OSS-v1 \\
NVIDIA & NemotronH-30B-A3B & 31.6B & 3.58B & 6 of 128 & Sophea-Nemo-3-Nano-v1 \\
NVIDIA & Nemotron-3.5-Lightning-30B-A3B & 31.6B & 3.58B & 6 of 128 & Sophea-Nemo-3.5-Lightning-v1 \\
\bottomrule
\end{tabular}
\caption{Total and routed-expert parameters counted from the released tensors; active-per-token is
non-routed $+$ $k/E$ of routed. Gpt-OSS ships MXFP4-packed experts, so its figures are
vendor-reported. NemotronH (Nano) and Nemotron-3.5-Lightning are \textbf{Mamba/MoE hybrids}
(Mamba, MoE and attention layers of 52; the state-space line of \citet{gu2023mamba,dao2024mamba2}
and the vendor's Nemotron-H report~\citep{nvidia2025nemotronh}); the other two are MoE transformers. The right-most column
names the fine-tuned Sophea reasoning release each base produces; all four SFT releases are
language-matched, and a fifth release, Sophea-Qwen3.6-v1.1, is the RLVR refinement of the Qwen row
(\S\ref{sec:rlvrresults}). Throughout, \emph{base} refers to
the vendor checkpoint and \emph{fine-tuned} to its Sophea reasoning model. Lightning is the
next-generation release in the Nemotron line; Nano and Lightning share the 31.6B / 6-of-128 routing
profile and are reported together as the Nemotron family (Nano is the same-generation member
of that family).}
\label{tab:models}
\end{table*}

\textbf{Total size differs by $1.7\times$ while active size differs by $11\%$.} All three cost
roughly the same to serve and spend that budget differently: Qwen routes 8 of 256 experts (3.1\% of
its parameters per token), Gpt-OSS 4 of 32 (12.5\%), NemotronH 6 of 128 while replacing most of its
attention with state-space layers. Comparing them on \emph{total} parameters would say Qwen is
$1.7\times$ the model that Gpt-OSS is; comparing on \emph{active} says they are the same size. We
report both and treat active as the meaningful axis, because it is what a deployment pays.

This also bounds what our results generalise to: MoE models are not evidence about dense ones
(Limitations). And MoE changes what LoRA attaches to.

\textbf{Adaptation.} LoRA~\citep{hu2022lora} at $r{=}32$, $\alpha{=}64$, one epoch, effective batch
$32$, learning rate $2\!\times\!10^{-4}$. All training, merging, and evaluation ran on a single
NVIDIA DGX~B200 node: $8\times$ B200 GPUs ($180$\,GB HBM3e each, $1.44$\,TB aggregate), which fits
every configuration in this paper without model parallelism beyond FSDP sharding, a deliberate
constraint, since a recipe for a low-resource language should be reproducible on one node. \emph{Expert stride 3} means adapters are placed on every
third MoE layer rather than all of them; the shared expert, which lies on every token's path, is
always included. The stride exists because expert count, not model size, sets the number of adapters:
where a model stores its experts as stacked parameters, PEFT attaches one adapter per stack
(28 for Qwen, 16 for Gpt-OSS), but where they are individually-materialised \texttt{nn.Linear}
modules it attaches one per expert: 4{,}188 separate adapter tensors on a model we profiled, which
costs step time in kernel launches and gradient all-reduces rather than in FLOPs. Fused-expert LoRA
also cannot be loaded back by \texttt{PeftModel.from\_pretrained} under our library versions, so
every Qwen checkpoint is merged into dense weights before evaluation.

\textbf{Corpus.} $118{,}092$ Greek rows in two halves of almost equal size.

The \emph{reasoning half} ($59{,}107$) carries an explicit trace in a separate field. It is
\textbf{$98.5\%$ synthetic}: questions and gold answers come from public English datasets, but the
traces are generated (see below). Only $894$ rows carry a trace we did not produce (the domain
mix is Table~\ref{tab:corpus}).

The \emph{direct half} ($58{,}985$) has no traces. It is the instruction corpus of
\textbf{Sophea-Titan-1}, a previously released Greek model, reused unchanged: Aya~\citep{singh2024aya}
($33.8\%$), a Greek instruction set ($25.8\%$), synthetic multi-turn dialogue ($10.1\%$), Greek QA
($7.7\%$) and twelve smaller sources, plus $\sim$$2{,}600$ rows of domain chain-of-thought (legal,
medical, finance, energy) and $50$ identity rows. It is $91\%$ Greek and $9\%$ English ($5{,}211$
rows), the English kept deliberately as replay against catastrophic forgetting. Reusing a corpus
that had already produced a working Greek model is why we did not initially suspect this half of
anything.

Its intended role is to keep the model's non-reasoning mode alive, and it does: \S\ref{sec:recipes}
shows that omitting it collapses the reasoning switch. It is also, on the same evidence, what costs
the model most of its accuracy and its answer-format compliance. That tension is the paper's one
replicated recipe finding.

\begin{table}[h]\centering\footnotesize
\setlength{\tabcolsep}{4pt}
\begin{tabular}{@{}lrr@{}}
\toprule
domain (reasoning half) & rows & share \\
\midrule
mathematics & 22{,}608 & 38.2\% \\
science & 11{,}064 & 18.7\% \\
world knowledge & 5{,}884 & 10.0\% \\
deductive logic & 5{,}151 & 8.7\% \\
medical & 4{,}898 & 8.3\% \\
commonsense & 3{,}904 & 6.6\% \\
reading comprehension & 3{,}828 & 6.5\% \\
physics & 1{,}770 & 3.0\% \\
\bottomrule
\end{tabular}
\caption{The reasoning half is math-heavy and thin on the axes we evaluate hardest: commonsense
is $6.6\%$ and logic $8.7\%$ of it.}
\label{tab:corpus}
\end{table}

\textbf{Where the traces come from.} Questions and gold answers are taken from public English
datasets and translated; the \emph{traces are generated}, not translated. Asking a translator to
render a chain of thought returns a tidy summary (which is the defect described next), so we
prompt an LLM to solve each question afresh in Greek and keep the trace only if its final answer
agrees with the gold. The generators are two frontier commercial models from a single family:
one for the mathematics slice (from
OpenR1-Math-220k questions) and a larger one for the logic, commonsense and regenerated-ECQA
slices); the remaining reasoning rows are carried unmodified from their sources
(Llama-Nemotron post-training science, medical-o1, Dolci). Every row carries a
\texttt{source} field recording its origin. Rows failing the gold check are discarded rather than
repaired (the answer-gating rule is STaR's~\citep{zelikman2022star}): a wrong trace teaches wrong
reasoning. Typical yield is $60$--$95\%$ depending on source.

\textbf{Trace genre, and how we measure it.} The property that distinguishes a useful trace from a
useless one is not correctness but \emph{structure}: whether the text shows reasoning happening
(a candidate tried, a flaw noticed, a correction) or merely justifies a conclusion already
reached~\citep{li2025structure,gandhi2025cognitive}. We score it as
\[
S \;=\; 0.40\,b \;+\; 0.25\,v \;+\; 0.25\,p \;+\; 0.10\,\ell
\]
where $b$ is the presence of backtracking markers, $v$ of verification, $p$ the fraction that is
flowing prose rather than a numbered list, and $\ell$ a length term. Backtracking dominates because
it is the property a write-up never has. $S$ is a \emph{corpus} diagnostic used to decide what to
train on; it is not one of the model metrics in \S\ref{sec:metrics}.

The distinction is not theoretical. A public Greek commonsense-explanation set (ECQA questions
translated into Greek with the original human-written justifications kept)
scores $S=0.27$ with \textbf{$0\%$} of its traces above $0.5$;
regenerating traces for the \emph{same questions} with the prompt above gives $S=0.68$ and $97\%$
above $0.5$. Our own first corpus was $72\%$ numbered write-ups, which is what sent us looking.

\textbf{Naming.} Arms are named for what actually distinguishes them (the training recipe and the
number of reasoning rows) rather than by internal version tags (Table~\ref{tab:naming}):

\begin{table}[h]\centering\footnotesize
\setlength{\tabcolsep}{4pt}
\begin{tabular}{@{}llr@{}}
\toprule
name & trained on & rows \\
\midrule
\textsc{Base} & nothing (the released model) & --- \\
\textsc{Reasoning} & reasoning rows only & 59{,}107 \\
\textsc{Subset} & a subset of the same rows & 15{,}607 \\
\textsc{Two-Phase} & reasoning rows, then $+$ the direct half & 59{,}107$+$ \\
\textsc{One-Phase} & one pass over both halves at once & 59{,}107$+$ \\
\bottomrule
\end{tabular}
\caption{\textsc{Two-Phase} continues the \emph{same adapter} from \textsc{Reasoning} onto the
hybrid mix; \textsc{One-Phase} never separates the two halves. The three recipes fail in different ways
(\S\ref{sec:recipes}).}
\label{tab:naming}
\end{table}

\textsc{Subset} is a draw of the same reasoning rows, originally selected by a trace-structure
score; \S\ref{sec:selection} shows that selection performs no better than sampling the same number
of rows at random, so we name it by what it is (a subset) rather than by a method the evidence
does not support.

\textbf{Benchmark.} $5{,}156$ Greek items: in mathematics, $250$ human-translated MGSM-style
GSM8K items (ILSP's \texttt{ilsp/mgsm\_greek}; MGSM itself~\citep{shi2023mgsm} contains no Greek,
so these are its $250$ GSM8K-test~\citep{cobbe2021gsm8k} problems human-translated by ILSP, the
Meltemi lab) plus our own machine-translated items from the rest of GSM8K test ($1{,}100$ of its $1{,}319$
items kept after number-preservation gating, minus $5$ duplicated in MGSM = $1{,}095$, $1{,}345$
math in total, $\S\ref{sec:lied}$);
Greek HellaSwag~\citep{zellers2019hellaswag} and WinoGrande~\citep{sakaguchi2021winogrande}
($3{,}267$ commonsense, machine-translated); and a decontaminated ProofWriter-el~\citep{tafjord2021proofwriter}
probe ($544$ logic, reported as macro-recall). Greedy decoding throughout.

\textbf{Scoring.} Every prompt requests a specific final line. We score \emph{that line} and report
\texttt{fallback\%}, the share of rows where the model never produced it (\S\ref{sec:lied}).

\textbf{Metrics.} Accuracy is one of six dimensions we report, not a summary of them: correctness,
language fidelity, reasoning budget, termination, reasoning steps and budget overrun. They are
defined in full in \S\ref{sec:metrics}, at the end, because the results are the argument and the
definitions are reference; a reader meeting \emph{budget overrun} or \emph{macro-recall} for the
first time in \S\ref{sec:quality} can turn there.

\textbf{Evaluation lanes.} The claims in this paper draw on five distinct instruments, and it
matters to keep them separate because they measure different things and are run under different
conditions. (i) The \textbf{Greek reasoning benchmark} above ($5{,}156$ items) is the home lane:
fidelity, accuracy, fallback, termination and budget behaviour all come from scoring full generated
traces on it, in both reasoning (\texttt{<think>}) and direct modes. (ii) The \textbf{English
control} ($1{,}100$ items: the gated English GSM8K originals behind the machine-translated Greek
math axis, \S\ref{sec:lied}; a separate held-out non-math probe accompanies it) asks whether the model still reasons in English when
the question is English; it is scored the same way, and it is what E1 (\S\ref{sec:langmatch}) is read
from. (iii) The \textbf{Titan-1 suite} (nine Greek plus five English NLU benchmarks, scored by
log-likelihood with no generation at all) measures \emph{general} ability for the forgetting
question (E2); it cannot see trace language, deliberately: it answers what SFT damaged, not how
the model writes. (iv) A \textbf{sentence-level switching probe} (E4b method: hand-labelling the
language of each sentence in $\sim$$150$ traces per condition) is what ``zero switches per 100
sentences'' is counted on; a per-trace ratio cannot see a mid-trace language island. (v) A
\textbf{Greek NLU benchmark} ($18$ tasks, $9{,}751$ items: NLI, sentiment, coreference, extractive
QA, cloze, machine translation, \ldots), generated in \emph{direct} mode with no reasoning
requested and scored strict/lenient, answers whether the non-reasoning mode still works (E3,
\S\ref{sec:plan}); its strict-vs-lenient split separates capability loss from format loss the same
way \texttt{fallback\%} does on the reasoning side. Where a number could have come from two
instruments, we name the lane next to the number.

\section{Metrics}\label{sec:metrics}

Accuracy is one of six dimensions, not a summary of them. Let an arm $a$ produce, on item $i$, a
reasoning trace $t_a(i)$ of length $w_a(i)$ words and a final answer.

\textbf{M1: Correctness.} Anchored accuracy: the requested answer line is parsed first, free-text
matching only if it is absent. Logic is reported as \emph{macro-recall} over its three classes,
never accuracy: the class distribution is $891/270/165$, so a majority-class answerer scores $67\%$
raw and $33\%$ macro. Accuracy on that axis measures prior-match, not reasoning. Per-class recall
shows how literal that is: \emph{every} arm, base included, is near-blind to \emph{Lathos}
(False, $14$--$16\%$ recall across the line) while \emph{Agnotsto} (Unknown) is answered at a
prior of $68.8\%$:
the third class works as a decision threshold between two priors, not as a reasoning output.
The same mechanism explains why this axis carries the seed swings of \S\ref{sec:variance}: a
single seed change moved one arm's \emph{Sosto} (True)-class recall $84.7 \to 17.3$ ($-67$ pp)
\emph{without} touching the other classes: a $\text{True} \to \text{Unknown}$ prior slide on the
same data, not a capability change. When a logic number moves, read the class recalls, not the mean.

\textbf{M2: Language fidelity.}
$g(t) = |{\rm GR}(t)| / (|{\rm GR}(t)| + |{\rm LA}(t)|)$, the Greek share of alphabetic characters
after stripping code and \LaTeX{} (otherwise technical traces read as Latin). We report the median
$g$ and the fraction of traces with $g \geq 0.9$.

\textbf{M3: Reasoning budget.} Median $w_a$, and \emph{words per correct answer}
$\sum_i w_a(i) / |\{i : a \text{ correct}\}|$, the quantity a deployment pays for. One confound
conditions every word- or token-based number in this paper: Greek costs $2.3$--$2.5\times$ the
tokens of English per word \emph{on all three families} (measured fertility: $2.32$, $2.39$,
$2.51\times$, see Limitations), so a uniformly-applied $\texttt{--max-new}$ budget is a
roughly $2.4\times$ tighter ceiling in Greek. Both M3 and M6 are reported in words, not tokens,
by construction: the metric exists precisely to hold the comparison fair across languages; the
tokenizer is the hidden hand behind them.

\textbf{M4: Termination.} Share of rows reaching the generation cap without a final answer.
Fully objective; no proxy.

\textbf{M5: Reasoning steps.} Count of \emph{intermediate conclusions}: trace sentences asserting
a numeric result or explicitly evaluating a named option. Not sentence count, which is
$w/12$ in disguise (\S\ref{sec:gate}).

\textbf{M6: Budget overrun.} With $f(i)$ the fraction of arms answering $i$ correctly and
$\tilde{w}(i)$ the median trace length across arms on $i$,
\[
{\rm OV}_a \;=\; \frac{\bigl|\{\,i : f(i) \geq 0.8 \;\wedge\; w_a(i) \geq 3\,\tilde{w}(i)\,\}\bigr|}
                      {\bigl|\{\,i : f(i) \geq 0.8\,\}\bigr|}
\]
(overspending on items nearly every model gets right, where difficulty cannot excuse it). Both
ingredients are model-independent: $f$ and $\tilde{w}$ are computed across all arms.

\begin{figure*}[t]
\centering
\includegraphics[width=\textwidth]{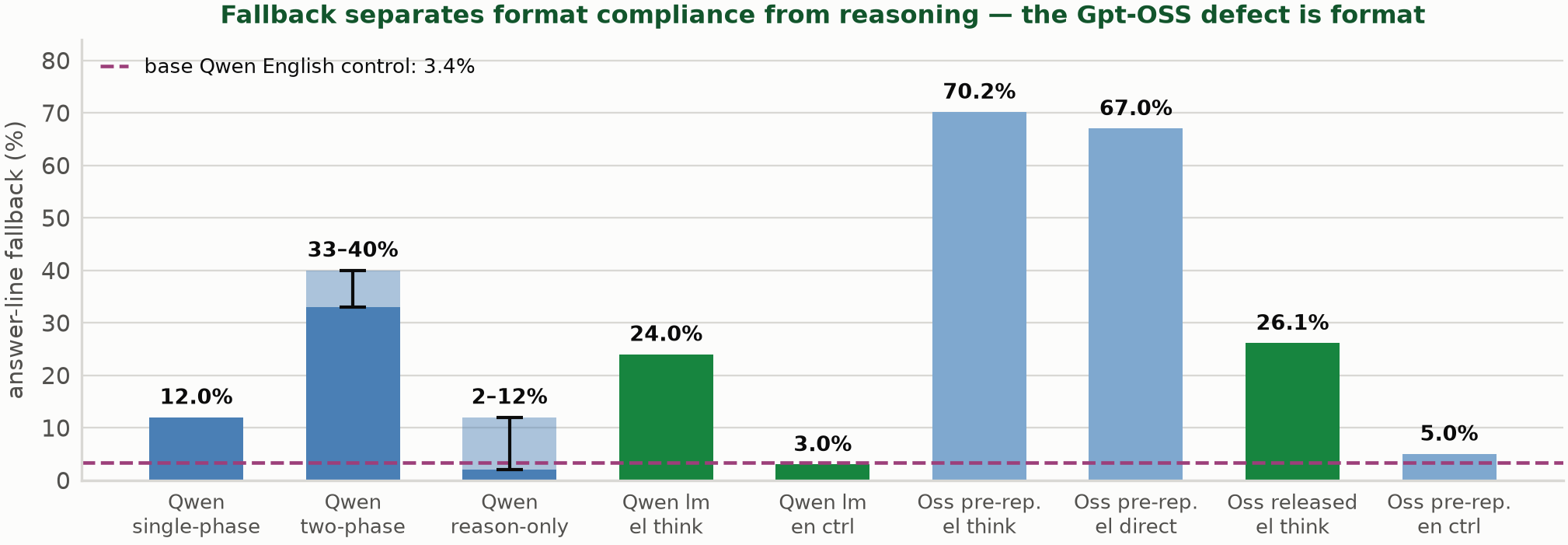}
\caption{\texttt{fallback\%} (the rate at which a model never emits the requested answer
line) by recipe and by language-matched checkpoint. It separates ``cannot reason'' from ``will not
answer in the requested form'': every recipe failure in the paper is visible here before it is
visible anywhere else. The Qwen recipes are the side experiment of \S\ref{sec:side}; the Gpt-OSS
language-matched arm's pre-repair Greek-lane rate (70\%) is the family defect at its largest
(its traces are present and on-language; the model simply does not close with the line
the scorer is told to read), and the released checkpoint's format-repair dose cuts it to $26\%$
(\S\ref{sec:plan}), still the highest of the releases. The dashed reference is a base model on
the English control ($3.4\%$),
showing the failure is not inherent to the benchmark.}
\label{fig:fallback}
\end{figure*}

Two diagnostics accompany these rather than scoring quality: \texttt{fallback\%} (M1's parse-failure
rate, which turns out to measure instruction-following; Figure~\ref{fig:fallback} maps it by
recipe and family) and \emph{switch integrity}, the share of
think-mode rows returning an empty trace when one was requested.

\section{The noise floor}\label{sec:variance}

\begin{figure*}[t]
\centering
\includegraphics[width=\textwidth]{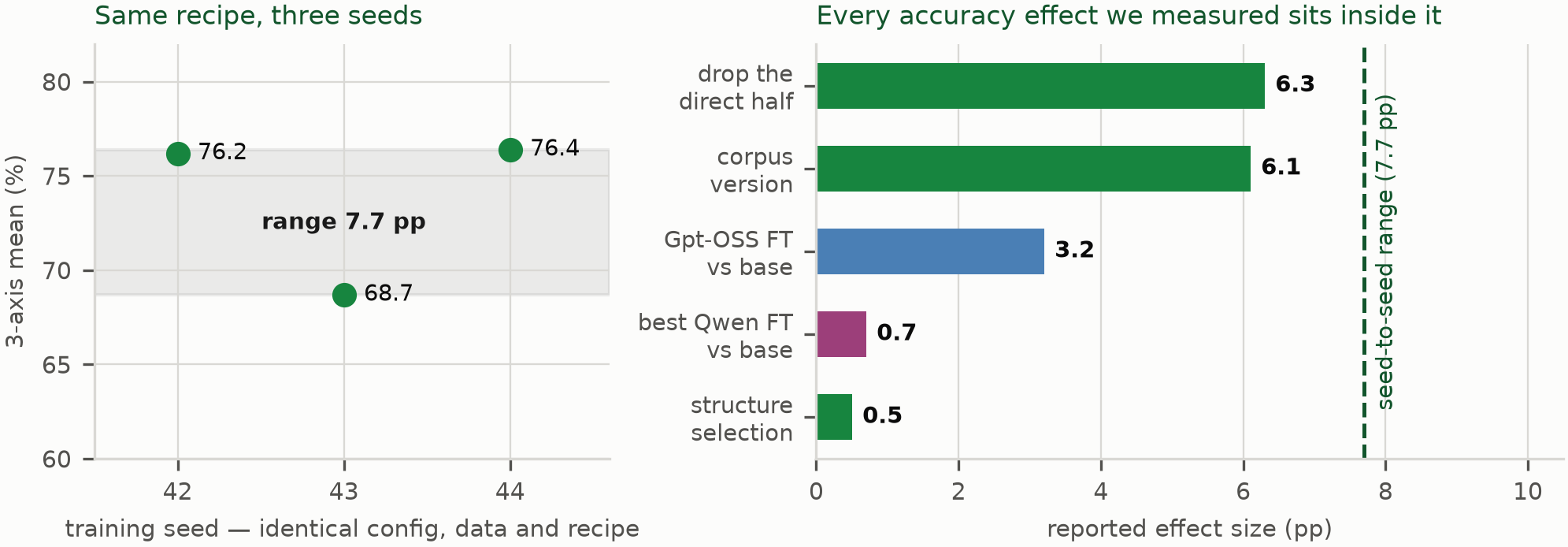}
\caption{Left: one configuration, three seeds, nothing else changed. Right: every accuracy effect we
measured over the project, against that range. All of them fit inside it.}
\label{fig:variance}
\end{figure*}

\subsection{Why vary the seed}

Every claim we had been making has the same shape: \emph{recipe A scored $x$, recipe B scored $y$,
therefore B is worse}. That inference carries a hidden assumption (that running recipe A twice
would return $x$ twice), and we had never tested it. In effect we were comparing interventions
against zero, having assumed the noise was zero.

Varying the seed tests exactly that assumption, and it is the control condition rather than an
additional experiment. The seed sets the LoRA $A$-matrix initialisation and the data shuffling
order; the corpus, hyperparameters, recipe and step count are identical, the benchmark is fixed and
decoding is greedy. Any spread that appears is therefore \emph{training}, and it is the distribution
of results produced by doing nothing at all: the null against which every measured effect has to
be read.

The closest analogy is calibrating an instrument: before claiming two objects differ in length, one
measures the same object twice to learn the ruler's precision. We had been reporting differences
finer than our ruler could resolve.

\textbf{This is not the error bar usually quoted.} The $\pm3.4$ pp figure that a benchmark of this size
implies is \emph{sampling} error: how much the score moves because 1{,}000 particular items were
drawn. It says nothing about how much the \emph{model} moves between runs, and training variance
turned out to be roughly twice as large. Papers reporting one number per configuration are
accounting for the smaller of the two sources. That training-run variance, not sampling variance,
is the operative error bar has been argued before~\citep{dodge2020show,bouthillier2021accounting,madaan2024variance};
\S\ref{sec:related} places this project on that line.

\subsection{What it cost, and what it bought}

We trained the same reasoning-only configuration three times, varying only the seed
(Figure~\ref{fig:variance}: the three runs on the left, every measured effect against the
resulting band on the right).

\begin{table}[h]\centering\small
\begin{tabular}{@{}lrrrr@{}}
\toprule
seed & mean & logic & fallback & trace Greek \\
\midrule
42 & 76.2 & 56.2 & 10\% & 1.00 \\
43 & \textbf{68.7} & \textbf{35.6} & \textbf{41\%} & 1.00 \\
44 & 76.4 & 54.0 & 3\% & 1.00 \\
\bottomrule
\end{tabular}
\caption{$\text{sd}=4.4$ pp, range $=7.7$ pp. Note the last column.}
\label{tab:seeds}
\end{table}

Two runs land near $76$ and one collapses to $68.7$, so this reads less like symmetric jitter than
an occasional failure mode: roughly one run in three lands $\sim$$8$ pp low, taking the logic axis
and instruction-following down together. Either reading supports the same conclusion. For a
difference $\Delta$ to be detected at the $5\%$ level with $90\%$ power against $\sigma \approx 4.4$ pp,
a two-sample comparison needs
\[
n \;\gtrsim\; 2\left(\frac{(z_{\alpha/2}+z_{\beta})\,\sigma}{\Delta}\right)^{2},
\qquad z_{\alpha/2}=1.96,\; z_{\beta}\approx 1.28,
\]
seeds per arm: about $11$ for our largest effect ($6.3$ pp) and about $40$ for the $3.2$ pp one. We
could not afford either, and that infeasibility is itself the finding.

Three runs also bound $\sigma$ itself only loosely: the $95\%$ chi-square interval on
$\hat\sigma = 4.4$ pp is $[2.3, 27.7]$ pp. The $7.7$ pp range is therefore a point estimate from
the one configuration and family it was measured on (Qwen); where this paper cites the band
against other families or recipes, it is an assumption carried across, not a measurement; the
qualitative conclusion (single-run deltas are unreliable) is what transfers, not the number.

\begin{figure*}[t]
\centering
\includegraphics[width=\textwidth]{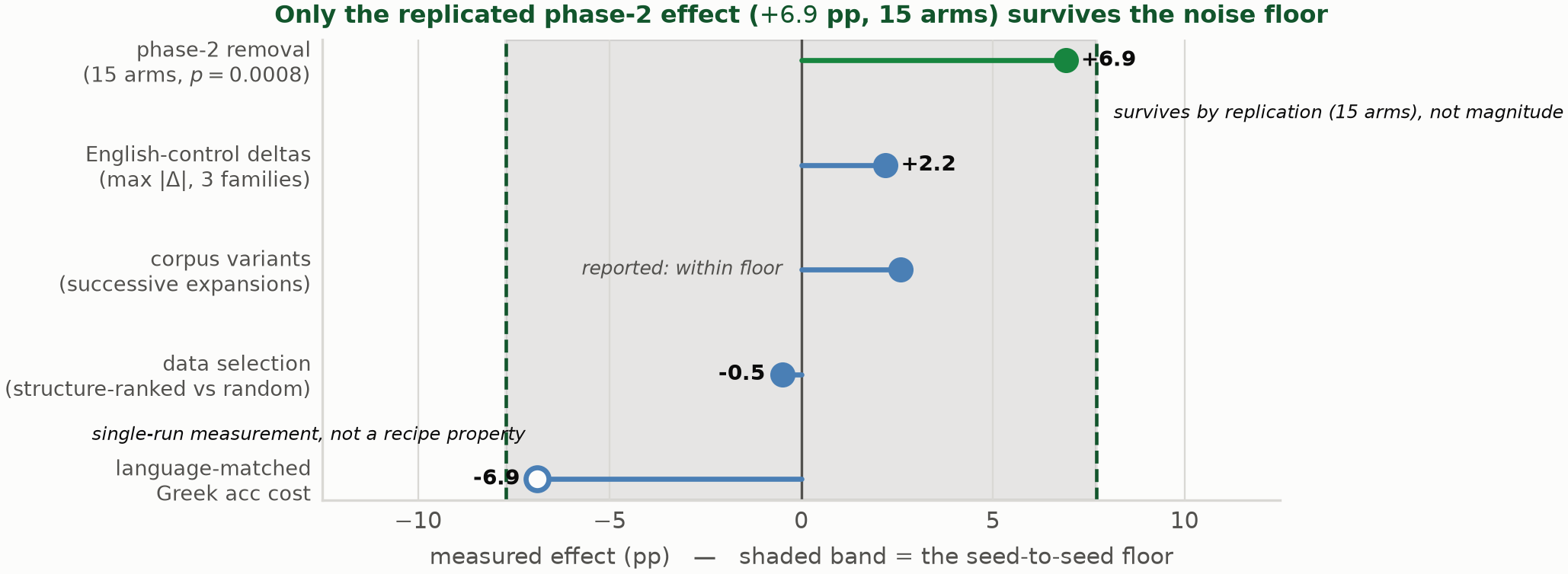}
\caption{Every accuracy effect we measured, against the $7.7$-point seed-to-seed band (shaded).
Attribute-level effects (data selection and corpus versions) sit inside the band we would
have been tempted to narrate. The only effect that survives is the replicated phase-2 comparison
of \S\ref{sec:side} ($+6.9$ pp, $15$ arms, permutation $p=0.0008$): it is inside the band
geometrically, but it is estimated across $15$ independently trained arms rather than read off a
single run, which is precisely the difference the floor teaches. The language-matched accuracy
cost ($-6.9$ pp) is drawn as an open (unfilled) marker on purpose: it is a single-run measurement inside the floor,
disclosed rather than interpreted, and not a recipe property. The equal magnitudes of the $+6.9$
and the $-6.9$ are a coincidence: different lanes, different arms.}
\label{fig:floor}
\end{figure*}

The control cost \textbf{one additional training run}. It invalidated five conclusions we had
already written down (Figure~\ref{fig:floor} plots every reported effect against the band), and
it redirected the paper: the same three seeds leave trace-Greek at
$1.00/1.00/1.00$ and trace length at $132/148/152$ words, so the behavioural dimensions are
\emph{stable} under precisely the perturbation that makes accuracy unusable. (One honest caveat:
trace-Greek sits at its ceiling, where low variance is partly definitional; the unsaturated
trace-length column, varying $132$--$152$ against an accuracy swing of $7.7$ pp, is the stronger
stability evidence.) We would not have
looked for them otherwise. Run this control before the ablations, not after them.

\keyfinding{\textbf{Finding 1.} Seed-to-seed variance on a 35B MoE LoRA fine-tune exceeds every data
or recipe intervention we tested, replicating the training-variance literature
\citep{dodge2020show,bouthillier2021accounting,madaan2024variance} at MoE-LoRA scale\footnote{With one
caveat we do not resolve: three of the four conditions are sparse MoE, so whether routing amplifies
seed variance relative to a dense model is an open question (Limitations).}. Single-run
accuracy deltas at this scale are noise. The qualitatively new observation is the asymmetry:
\emph{the language and budget dimensions do not move across the same three seeds}: trace-Greek
at $1.00/1.00/1.00$ (Table~\ref{tab:seeds}, last column) and trace length within $132$--$152$
words. That is what lets the rest of the paper measure anything. The asymmetry is not
universal: \texttt{fallback\%} swings $3$--$41\%$ across the same seeds (the same table's third
column), moving \emph{with} accuracy rather than against it (\S\ref{sec:side}).}

\subsection{Data selection does nothing}\label{sec:selection}

The clearest single demonstration is a control we should have run first. \textsc{Subset} was
built by scoring every reasoning row for trace structure and keeping the top $15{,}607$ (mean
structure $0.836$). We then trained the identical recipe on $15{,}607$ rows drawn \emph{uniformly at
random} from the same pool (mean structure $0.587$; Table~\ref{tab:selection}):

\begin{table}[h]\centering\footnotesize
\begin{tabular}{@{}lrr@{}}
\toprule
phase-1 rows & selection & mean \\
\midrule
62{,}562 & all & 69.5 \\
15{,}607 & top-by-structure & 69.5 \\
15{,}607 & uniform random & 69.0 \\
\bottomrule
\end{tabular}
\caption{Selection buys $-0.5$ pp ($0.31\sigma$); a $4\times$ smaller pool costs nothing. The
$62{,}562$ is the reasoning count of the pre-gate \emph{pool} the experiment was run against
($62{,}562$ reasoning, $60{,}214$ direct); the
$59{,}107$ quoted everywhere else is the post-gate \emph{train} half
($59{,}107$ reasoning, $58{,}985$ direct), so the table's
\emph{all} row and the paper's reasoning-half count differ by the $3{,}455$ rows the
decontamination/dedup gate and the val reserve removed.}
\label{tab:selection}
\end{table}

The pre-registered prediction was that selection would win by more than 2 pp. It is worth being
explicit that an ordering across our corpus versions had earlier appeared to show ``smaller is
better''; that ordering was confounded with which corpus each subset came from, and does not
survive this control. \citet{xia2024random} report the same null at scale.

\section{What actually changed}\label{sec:quality}

\begin{figure*}[t]
\centering
\includegraphics[width=\textwidth]{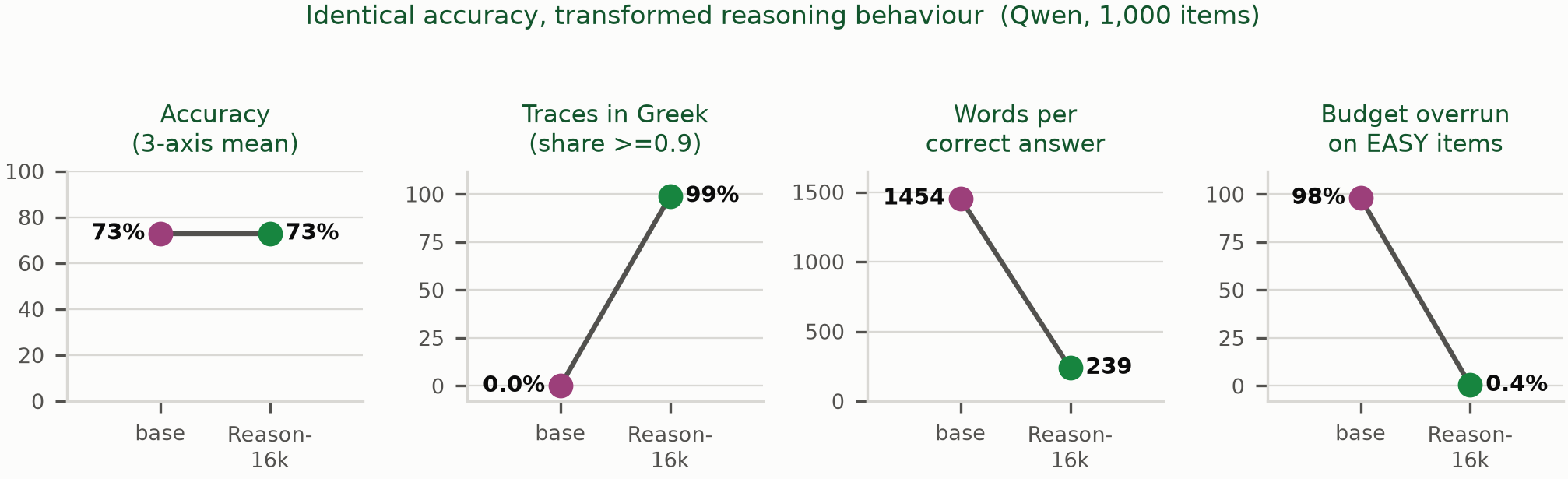}
\caption{The same two checkpoints on four dimensions, the base and the structure-selected
reasoning-only arm (\textsc{Subset}, the ``Reason-16k'' of Figure~\ref{fig:domain}), on the
$1{,}000$-item probe lane. Only the first is accuracy.}
\label{fig:quality}
\end{figure*}

\subsection{Language fidelity}

Given a Greek question, the base models never reason in \emph{Greek}, and mostly not in pure
English either. Zero of $1{,}000$ traces reach a
Greek character ratio of $0.9$ (rule-of-three $95\%$ upper bound $0.3\%$; the zero holds again on
all $5{,}156$ items, Figure~\ref{fig:langmatrix}), the median ratio is $0.33$
(English-scaffolded mixed script, only $4.2\%$ purely English), and reading them confirms the
English scaffolding (``\texttt{We need
to answer a multiple-choice question in Greek\ldots}''). Every fine-tune we trained reasons in
Greek on $97.4$--$98.7\%$
of measured traces (the four SFT releases at $97.4$--$98.1\%$,
Table~\ref{fig:families}; the one-directional Qwen arm at $98.7$; the RLVR release at $98.27$,
\S\ref{sec:rlvrresults}).

In our study the property first arrived \emph{overcorrected}: the one-directional fine-tune,
instructed explicitly to reason in English, produced a non-English trace on $1000/1000$ items:
SFT moved something no bare instruction we tried could reach (few-shot untested; Limitations),
and moved it too far, since a model that cannot be asked to think in English has lost a
capability. The correct target is matching the \emph{question}; retraining on language-matched
pairs reaches it (Greek fidelity $97.98\%$ ($5{,}052/5{,}156$), English traces on $100\%$ of
the $1{,}900$ paired English items, zero in-question switches either direction), with the full
account, including the partial re-opening of the instructed override on two of the four released
checkpoints, in \S\ref{sec:langmatch} and \S\ref{sec:override}.

We test whether the recipe is Qwen-specific: trained on Gpt-OSS-20B, the identical
language-matched mix measures $97.79\%$ Greek fidelity pre-repair and $98.14\%$ on the released,
format-repaired checkpoint (both within one binomial standard error of the Qwen release),
with zero switches in both modes and $100\%$ English-control compliance
(Table~\ref{tab:encontrol}); Nemotron-3.5-Lightning, trained
on the same mix, measures $98.06\%$ fidelity, zero switches and $100\%$ English-control compliance
under the identical probe (Table~\ref{fig:families}), with general-ability retention measured
against its own base: $+1.7$ points Greek macro and $-1.1$ English on the Titan-1 suite
(\S\ref{sec:plan}, Table~\ref{tab:forgetbench}); it is released as a third matched checkpoint,
not a full replication of the two. The
Gpt-OSS arm does show one
family-specific cost: its answer-channel leak is $9.1\%$ pre-repair and $10.3\%$ on the released
checkpoint, against Qwen's $3.5\%$, so the
trace-register boundary that \S\ref{sec:metrics} measures holds at different tightness per family.

The Gpt-OSS arm also separates the two failure modes this paper keeps apart. Before its format
repair, its Greek-lane anchored accuracy reads $56.2$, with $70.2\%$ of rows never emitting the
requested answer line,
so the score is almost entirely a \emph{format} floor, not a reasoning one (\S\ref{sec:metrics}):
the reasoning is present and on-language; the model does not close with the line the scorer reads.
Both language-matched checkpoints pay a format tax: Qwen's $\texttt{fallback\%}$ is $24$ on the
Greek lane (and $3$ on the English lane); the released checkpoint answers in the requested form
on three of four Greek items. Gpt-OSS's is $70$/$67$ (think/direct) before the repair, with $5$
on English; the released, format-repaired checkpoint pays $26$ on the Greek think lane
(\S\ref{sec:plan}), still the highest of the releases.
The fidelity result above is independent of this metric: it is measured on the trace, where the
released checkpoints are on-language at $98.0$--$98.1\%$.

Reading the same releases per domain sharpens that format-versus-capability reading
(Table~\ref{tab:domain}). The defect the Gpt-OSS arm pays is not uniform: before repair its
fallback was $\textbf{35\%}$ on math against $\textbf{83\%}$ on commonsense and $\textbf{77\%}$ on
logic: the loss lives in the open-form domains, while math's short numeric answer is
format-robust. That is the shape a format defect takes, not the shape a capability loss takes.
Repair rescues the worst domain first (commonsense $83{\to}26\%$ and logic $77{\to}51\%$
fallback, math $35{\to}17\%$), and
accuracy follows. Two smaller readings. Logic is the hardest axis for \emph{every} arm (base
macro-recall $32$--$43\%$ across the two base rows), so its low scores are a floor, not a
regression. And a correction this
revision owes the reader: an earlier draft read the Qwen release as \emph{gaining} on commonsense
($63.2{\to}75.8$), but the $63.2$ comparator in Table~\ref{tab:domain} is the \emph{Gpt-OSS}
base; against its own base's $82.4$ on the same lane, the Qwen release is $-6.6$ on commonsense, and
no fine-tune beats its own base on any axis of Table~\ref{tab:domain}. The Qwen release's logic
fallback is $46\%$, and the repaired Gpt-OSS release's higher still ($51\%$).

\begin{table*}[t]\centering\small
\setlength{\tabcolsep}{6pt}
\begin{tabular}{@{}lccc|ccc@{}}
\toprule
& \multicolumn{3}{c|}{anchored accuracy} & \multicolumn{3}{c}{fallback \%} \\
arm & math & cs & logic & math & cs & logic \\[-1pt]
 & \multicolumn{3}{c|}{$\uparrow$ higher is better} & \multicolumn{3}{c}{$\downarrow$ lower is better} \\
\midrule
Qwen base (Greek think) & 92.9 & 82.4 & 42.6 & 3.3 & 0.5 & 0.2 \\
\quad Sophea-Qwen3.6-v1 & 83.3 & 75.8 & 37.3 & 18.2 & 22.9 & 46.3 \\
\midrule
Gpt-OSS base (Greek think) & 90.7 & 63.2 & 32.2 & 2.4 & 5.2 & 0.7 \\
\quad \emph{before repair (reference)} & 74.6 & 53.7 & 25.2 & 34.9 & \emph{83.5} & \emph{77.4} \\
\quad Sophea-OSS-v1 (repaired) & 78.6 & 53.3 & 26.5 & 17.2 & 25.7 & \textbf{50.9} \\
\midrule
NemotronH base (Greek think) & 86.6 & 49.2 & 26.8 & 8.3 & 6.2 & 6.6 \\
\quad Sophea-Nemo-3-Nano-v1 & 71.5 & 39.7 & 9.8 & 1.7 & 1.8 & \textbf{39.9} \\
\midrule
Nemotron-3.5 base (Greek think) & 76.9 & 46.3 & 25.0 & 1.4 & 1.9 & 4.4 \\
\quad Sophea-Nemo-3.5-Lightning-v1 & 75.6 & 40.9 & 23.8 & 16.1 & 14.7 & 5.3 \\
\bottomrule
\end{tabular}
\caption{Per-domain results on the Greek think lane; here and in every table, $\uparrow$ marks metrics where higher is better and $\downarrow$ where lower is better (math = the two math slices, commonsense =
HellaSwag+Winogrande, logic = ProofWriter macro-recall). Rows are grouped by family, each release
indented under its own base, so every legitimate accuracy delta is a within-block comparison and
the cross-base reading (the one an earlier draft of this paper fell for) is structurally
discouraged. Every release now carries its own base row: the NemotronH base dump is the same
vintage as the Qwen base dump, and the Nemotron-3.5 base lane, absent until this revision, was
generated with the identical lane settings and scored with the same scorer as every other row.
The same base is also measured on the NLU retention suite (Table~\ref{tab:forgetbench}, Greek
macro $57.5$), a direct-mode instrument whose numbers are not comparable to this think-lane
table. The Nano release also appears in the retention suite (Table~\ref{tab:forgetting}),
and its fallback shape is the mirror of Gpt-OSS's: negligible on math and commonsense
($1.7$/$1.8\%$) but $39.9\%$ on logic, so its $9.8$ logic macro-recall is substantially a format
floor on the one domain where its answer form breaks. Accuracy and fallback
move together: every release keeps the base's domain ordering, and the Sophea-OSS format-repair
dose pulls the commonsense fallback back toward math's level ($83.5{\to}25.7$) while cutting
logic's by a third ($77.4{\to}50.9$). The pre-repair row is
reference only; the release is the repaired arm. Conditioning on rows that emit the requested
answer line (excluding the fallback-scored path) shrinks the apparent deficits sharply: the
Qwen release reads $95.8/78.8/39.5$ against its base's $96.0/82.8/42.7$, and the same condition
makes Gpt-OSS pre-repair $74.6\to92.2$ / $53.7\to60.2$ / $25.2\to36.5$ and Sophea-OSS-v1
$78.6\to94.4$ / $53.3\to58.2$ / $26.5\to30.0$ (math/commonsense/logic), so a substantial share
of every raw gap is format compliance, not reasoning.}
\label{tab:domain}
\end{table*}

One more observational claim we believed and then had to withdraw belongs here, because it is the
question every reader asks next: \emph{is Greek reasoning worse reasoning?} An early run of the
fine-tune on Greek items scored $-17.5$ pp against the same items in English, which reads
seductively as ``the model thinks better in English.'' The controlled version separates the two:
run the \emph{same} items, force the \emph{same} answer language, vary only the trace language
(\S\ref{sec:lied}'s answer-format control is the same trick in miniature). The language effect
collapses to $+1.4$ pp ($0.65\sigma$): statistically nothing. The $-17.5$ pp was selection
(a different, easier question mix in the English lane) expressing itself as a language effect,
and it is the largest retelling-of-a-difference the project ever wrote into a slide before the
control.

\begin{table}[!htb]\centering\footnotesize
\setlength{\tabcolsep}{2.2pt}
\begin{tabular}{@{}lrrrr@{}}
\toprule
 & Greek fid. & EN ctrl & switches & leak \\
arm & (\%)$\uparrow$ & (\%)$\uparrow$ & /100 sent.$\downarrow$ & (\%)$\downarrow$ \\
\midrule
Qwen base & 0.0 & 100 & 16.4 & 15.6 \\
Qwen lang-matched & 98.0 & 100 & 0.0 & 3.5 \\
Qwen one-dir.$^{\dagger}$ & 98.7 & 99.3$^{\S}$ & 0.0 & 1.5 \\
Gpt-OSS base & 0.0 & 100 & 9.7 & 0.5 \\
Gpt-OSS lang-matched & 98.1 & 100$^{\ddagger}$ & 0.0 & 10.3 \\
Nemotron-3.5 lang-matched & 98.1 & 100 & 0.0 & 5.2 \\
NemotronH lang-matched & 97.4 & 100 & 0.0 & \textbf{0.0} \\
NemotronH one-dir.$^{\dagger}$ & 98.4 & 100 & 0.0 & 11.3 \\
\bottomrule
\end{tabular}
\caption{Language-matching across the three families: all four released checkpoints, the two
measurable bases, and the one-directional reference arm. \emph{Greek fid.}: Greek-trace fidelity
on the Greek benchmark; both bases sit at exactly $0.0$ ($0/5{,}156$ traces reach ratio $0.9$),
and all four language-matched checkpoints land at the same level ($97.4$--$98.1\%$).
\emph{EN ctrl}: English-control compliance, $100\%$ for every arm. \emph{Switches}: in-question
language switches per $100$ sentences on the Greek lane; the bases switch $16.4$ (Qwen) and $9.7$
(Gpt-OSS) times, every fine-tuned arm (the one-directional reference included) zero. The base fidelity and switch cells are one
instrument run: both base dumps rescored with the current scorer on the same day (an earlier
draft printed $19.0$ for the Qwen base from an earlier dump vintage).
\emph{Leak}: answer-channel leak, the one place families differ ($10.3\%$ Sophea-OSS-v1,
$5.2\%$ Lightning, $3.5\%$ Qwen, $0.0\%$ Nano, the only arm at zero; base leak is measured on
the English control, fine-tuned leak on the Greek think lane). $^{\dagger}$Arms of the
earlier one-directional recipe, shown for reference (a different recipe, not releases); the Qwen
one-directional row is the recipe-evolution comparison: it locks against explicit language
\emph{instructions} (Table~\ref{tab:forget}), not against English questions.
$^{\S}$Measured on the $1{,}900$-item paired probe ($99.3\%$ English, $0.0\%$ Greek, $0.7\%$ too
short to score); this arm was never run on the $1{,}100$-item control the other EN-ctrl cells
use. $^{\ddagger}$Measured on the pre-repair language-matched arm: the repair dose adds $2\%$
trace-less answer-format rows and does not touch trace language, but the released arm's own
English control has not been generatively re-run. The recipe generalises to every family tried; the leak's size does
not.}
\label{fig:families}
\end{table}

\subsection{Reasoning budget: words fall everywhere it matters, tokens change sign by family}

\begin{figure}[h]
\centering
\includegraphics[width=\columnwidth]{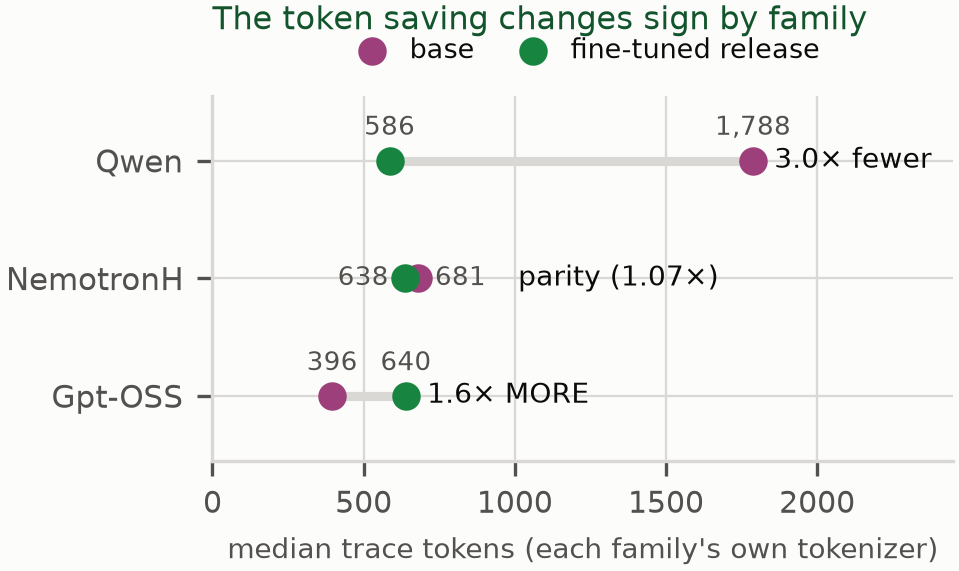}
\caption{Median trace length in \emph{tokens} (each family's own tokenizer, $5{,}156$ matched
items), base against the fine-tuned release. The word-level shortening must repay Greek's
$2.3$--$2.5\times$ token fertility before it becomes a serving saving: Qwen's does, NemotronH's
breaks even, Gpt-OSS's does not.}
\label{fig:tokens}
\end{figure}

At identical item-pooled accuracy ($72.9$ vs $72.9$ on the same $1{,}000$-item probe; the
three-axis strict mean there is $77.2$), the base spends a median $1{,}010$ words per trace
and $1{,}454$ words per \emph{correct} answer; the reasoning-only fine-tune spends $150$ and $239$
(Figure~\ref{fig:quality} sets this beside the flat accuracy panel).
In \emph{words}, our length unit throughout (\S\ref{sec:metrics}), the base-to-fine-tune ratio
is $3.7\times$ for the Qwen language-matched release and $1.5\times$ for the NemotronH fine-tune
(the earlier one-directional arms are more favourable still: $4.5$--$5.1\times$ Qwen, $1.4\times$
Gpt-OSS). But words are not what a deployment pays, and Greek costs $2.3$--$2.5\times$ the tokens
per word that the bases' English traces do (the measured fertility ratios of \S\ref{sec:metrics}). Retokenizing every trace on the full benchmark with
each family's own tokenizer, the saving changes sign by family: the Qwen release spends
$\mathbf{3.0\times}$ \emph{fewer} tokens than its base (median $586$ vs $1{,}788$), the NemotronH
fine-tune sits at \textbf{parity} ($638$ vs $681$, $1.07\times$), and the Gpt-OSS release spends
$\mathbf{1.6\times}$ \emph{more} ($640$ vs $396$): its traces are already slightly longer in
words than its terse base's, and the fertility gap widens the difference. We therefore split the
claim: the reduction in reasoning \emph{effort} (words) is real wherever the recipe shortens
traces, but the \emph{serving-cost} saving survives translation into tokens only on Qwen; on
Gpt-OSS the Greek fine-tune is more expensive to serve than its English-reasoning base
(Figure~\ref{fig:tokens}).

We state this as a marginal claim only. Conditioning on trace length, the base is at least as
accurate as the fine-tune, and \emph{within every arm longer traces are less accurate} (base $82.5\%$
under 400 words vs $72.3\%$ over). Length marks items a model is struggling with, not effort that
pays off. The fine-tune's advantage is that it does not enter the long unproductive regime, not
that its tokens are worth more. Both stratifications are endogenous, so no causal claim about length
is available from this design.

\subsection{Budget discipline}

\begin{figure}[h]
\centering
\includegraphics[width=\columnwidth]{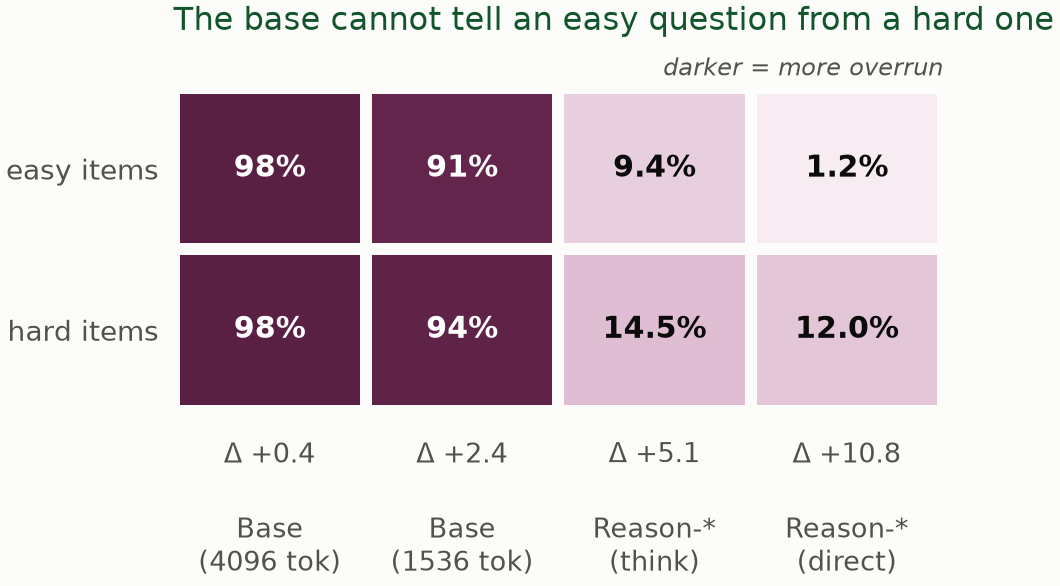}
\caption{Overrun $=$ trace $\geq 3\times$ the median trace \emph{for that item} across arms. Easy $=$
at least $80\%$ of arms answer correctly. Cells pool the arms of each column; hard-item think-mode
rates are compressed by generation-cap truncation (\S\ref{sec:quality}), so the easy row carries
the section's claims.}
\label{fig:overrun}
\end{figure}

Measuring degenerate looping directly proved impossible: looping is what \emph{makes} a trace long,
so every repetition metric we tried was a length proxy (\S\ref{sec:lied}). The measurable question is
whether a model spends far more than a given item requires. Let $f(i)$ be the fraction of arms
answering item $i$ correctly and $\tilde{w}(i)$ the median trace length across arms on that item; an
arm overruns on $i$ when $w(i) \geq 3\,\tilde{w}(i)$. Conditioning on the item supplies the control a
raw rate lacks.

The base overruns on \textbf{98.0\%} of easy items, against $0.2$--$12\%$ for every fine-tune
(Figure~\ref{fig:overrun}). And
its rate is identical on easy and hard items ($98.0$ vs $98.4$): it has no representation of question
difficulty. Fine-tuned arms range $1.9$--$60\times$ between the two in \emph{direct} mode, spending
more only when the item is harder; think-mode ratios are compressed by generation-cap truncation
on hard items, which is why the easy-item \emph{rate}, unaffected by the cap, is the figure this
section reports.

\keyfinding{\textbf{Finding 2.} We find that at equal accuracy the fine-tune reasons in Greek rather
than English and adapts its budget to question difficulty, but the token cost of doing so is
family-dependent in \emph{sign}: $3\times$ fewer tokens on Qwen, parity on NemotronH, $1.6\times$
more on Gpt-OSS, because Greek's $2.3$--$2.5\times$ token fertility must be repaid by shortening
traces (\S\ref{sec:quality}). None of these appear in a benchmark mean, and none move across
seeds.}

\section{Losses that looked like forgetting}\label{sec:forget}

If SFT installs Greek reasoning, what does it remove? The answer sorts into three buckets, two of
which are measured here, and the third is the one readers most often conflate with forgetting:
the language lock. That is a capability keyed to the training direction, not a deletion, and its
full treatment lives with the pre-registered E1 design in \S\ref{sec:langmatch}, including
Table~\ref{tab:forget}. For general ability itself, the forgetting question is answered on the
Titan-1 suite in that same section (Table~\ref{tab:forgetting}): flat in both languages on two
families, with a $-3.2$-point Greek residual on the format-repaired Gpt-OSS release. This section
carries the two remaining appearances of forgetting, both of which dissolve
on contact with the right control.

\textbf{(a) Format compliance degrades, and the recipe decides how much.} Failure to emit the
requested answer line rises from $12\%$ (single-phase) to $33$--$40\%$ (two-phase) and stays low
$2$--$12\%$ (reasoning-only, with seed noise widening that to $2$--$41\%$ as disclosed in
Table~\ref{tab:recipes}). The forgotten capability is format compliance, and \S\ref{sec:recipes}
localises it to the second phase. It is a recipe choice with a measured mechanism, not
forgetting: the language-matched checkpoints of \S\ref{sec:quality} pay it too, at $24\%$ (Qwen)
and $26\%$ (the format-repaired Gpt-OSS release; $70\%$ before its repair) on the Greek lane.

\textbf{(b) The commonsense loss is not forgetting.} A $-3.0$ pp commonsense deficit appeared in
$12$ of $12$ fine-tuned arms and looked like textbook catastrophic forgetting. Re-asking the same
items in a constrained answer format reverses it to $+1.7$ pp (\S\ref{sec:lied}). The capability was
never lost; only its expression under long-form generation changed. We report this because the
negative result is the more useful one: an apparent forgetting effect that survives twelve arms can
still be an artifact of how the answer was elicited.

\textbf{Register control and grammaticality: not lost, and mostly gained.} Two judge-based probes
close the forgetting picture from the fluency side, run identically on every release and its own
base (an LLM judge at temperature $0$; $n{=}46$ register items and $n{=}58$ morphosyntax items,
small enough that we report counts, not percentages). \emph{Register control} (produce the
requested formal or informal register) is never lost: the two weakest bases gain it
(Nemotron-3.5, $37/46 \to 43/46$; NemotronH, $40/46 \to 44/45$), Gpt-OSS is flat
($38/45 \to 39/46$), and Qwen dips within small-$n$ noise ($45/46 \to 42/46$).
\emph{Grammatical correctness} (agreement, clitics, and related morphosyntax) improves on every
family, and improves most where the base is weakest: NemotronH $13/58 \to 27/57$,
Nemotron-3.5 $12/58 \to 29/58$, Gpt-OSS $32/57 \to 40/58$, Qwen $41/58 \to 42/58$. One
instrument note in this section's own spirit: the first-pass Gpt-OSS numbers were an artifact of
a response-splitting bug that prefixed a stray channel marker to otherwise correct answers; the
judge failed them for the prefix, the splitter was fixed, and the fine-tune's dumps were
re-judged (the corrected numbers are the ones above, and the direction of the finding reversed
from loss to gain).

\keyfinding{\textbf{Finding 3.} We find that almost nothing the fine-tune appeared to forget was
actually forgotten. The language lock is conditional on training direction, not a deletion (full
treatment: \S\ref{sec:plan}); general ability is flat on two families, with a $-3.2$-point Greek
residual on the third after its format repair (Table~\ref{tab:forgetting}); the commonsense
regression was a format artifact; and the one genuine loss (answer-format compliance) is
recipe-localised and disclosed for the release checkpoints.\par\vspace{2pt}
On the fluency side the sign flips
outright: register control is retained or gained on every family, and judged grammaticality
improves on all four, most where the base is weakest. What a benchmark calls forgetting is,
in this project, almost always something else. Flat macros are also partly the expected property
of a LoRA adapter rather than of our data~\citep{biderman2024lora}; \S\ref{sec:plan} carries that
scoping in full, and the Limitations section keeps it LoRA-conditional on purpose.}

\section{A side experiment: three recipes, and what the second phase costs}\label{sec:side}
\label{sec:recipes}

This section stands apart from the paper's main line. The question it answers (\emph{how much of
the training corpus should be non-reasoning, and in what order should the halves be shown}) was
settled for our release long before the language-matching result existed, and none of the main
claims depends on it. It stays in the paper for two reasons: it supplies the mechanism for our
answer-format failure mode (\S\ref{sec:forget}b), and it contains the one accuracy effect in the
project that survived the noise floor, a result we would have to explain the absence of if we
omitted it. Read it as a controlled detour, not as a load-bearing section.

\subsection{The three recipes, and the trade they make}

The corpus has two halves: reasoning rows carrying explicit traces, and $\sim$$59$k non-reasoning
conversational rows. Three ways to use them, and each fails differently.

\begin{table}[h]\centering\footnotesize
\setlength{\tabcolsep}{4pt}
\begin{tabular}{@{}lccc@{}}
\toprule
recipe & mean & empty & fallback \\
\midrule
\textsc{One-Phase} & 66.1 & \textbf{23.6\%} & 12\% \\
\textsc{Two-Phase} & 64.4--69.9 & 0.0--1.3\% & \textbf{33--40\%} \\
reasoning-only & \textbf{73.6} & 0.0\% & \textbf{2--12\%} \\
\bottomrule
\end{tabular}
\caption{\emph{single-phase} = one pass over both halves; \emph{two-phase} = reasoning-only then
hybrid; \emph{reasoning-only} = the reasoning half alone. Empty-trace rate is measured in think mode,
\texttt{fallback\%} in direct mode; means are over all arms of each recipe. Fallback ranges are
over the individual arms of each recipe (2--12\% across the three
reasoning-only arms; 33--40\% across the two-phase arms); the three seed
replicates of \S\ref{sec:variance} widen it further (3--41\% across seeds 42--44,
all three of which are among the six phase-1-only arms above), so the fuller picture is
2--41\% when seed noise is admitted.}
\label{tab:recipes}
\end{table}

\textbf{In our runs, single-phase SFT collapsed the reasoning switch.} Trained in one pass over a corpus that is
half non-reasoning, the model learns to answer directly regardless of the flag: asked to think, it
returns an \emph{empty} trace on $23.6\%$ of items. This is the cleanest causal result in the line
and it is coverage-independent: it counts empty traces, not answers.

\textbf{Two-phase fixes the switch and breaks something else.} Training reasoning-only first, then
continuing the \emph{same adapter} on the hybrid mix, drives empty traces to $0$--$1.3\%$. But
answer-format compliance degrades from $12\%$ to $33$--$40\%$ fallback, and accuracy does not
improve. The second phase repairs the mode switch at the cost of instruction-following.

\textbf{Phase-1-only avoids both, and gives up the direct mode.} Never showing the model the
non-reasoning half yields the best accuracy \emph{and} the best compliance. At the time this
experiment ran, the cost was architectural rather than measured: such a model has only ever been
trained to reason, so it has no separately trained direct mode to collapse. E3
(\S\ref{sec:plan}) later measured that cost directly, and found none detectable on direct-mode
NLU.

\subsection{The one accuracy effect that survives the floor}

\begin{figure}[h]
\centering
\includegraphics[width=\columnwidth]{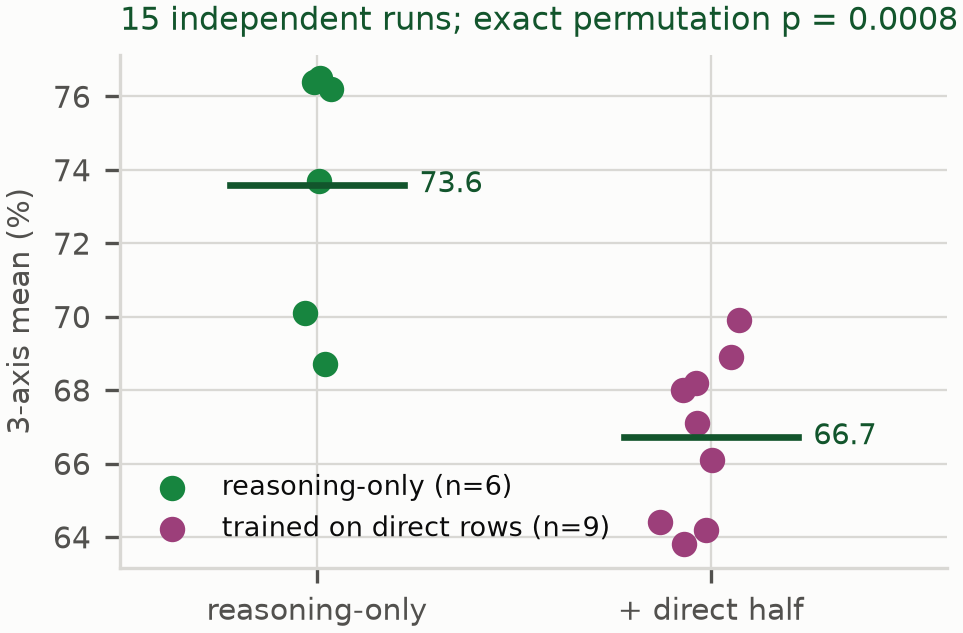}
\caption{Each point is an independently trained arm. Corpus version varies within both groups.}
\label{fig:phase}
\end{figure}

A single reasoning-only versus two-phase comparison ($+6.3$ pp) sits inside the noise floor and cannot
be claimed. But the distinction separates \emph{fifteen} independently trained arms
(Figure~\ref{fig:phase}), and across runs
that is the correct unit of analysis: reasoning-only averages $73.6$ ($n{=}6$) against $66.7$ ($n{=}9$),
with $52$ of $54$ pairwise comparisons favouring it and an exact permutation $p = 0.0008$. On the
decontaminated benchmark the effect is $+5.3$ pp ($p = 0.0164$), and on the $5{,}156$-item benchmark
reasoning-only wins \emph{every axis} ($+5.4$ pp mean).

One confound this design cannot fully exclude: the arms are the project's historical runs, and
corpus version varies within both groups rather than being balanced across them. Stratifying the
permutation by corpus version leaves only three strata with members on both sides and $16$ valid
permutations in total: the observed split ranks second of the sixteen ($p = 0.125$, attainable
floor $0.0625$), so the stratified test is supportive but has almost no resolution. The
$p = 0.0008$ figure assumes exchangeability
across corpus versions; the fact that four versions appear on both sides argues for that assumption
qualitatively, but a designed, version-balanced replication is what would settle it.

The mechanism is visible in \texttt{fallback\%}: reasoning-only arms fail to emit the requested answer
line on $2$--$12\%$ of items, two-phase arms on $33$--$40\%$. The wider range in the seed replicates
reasoning-only alone (3--41\% across the three seeds of \S\ref{sec:variance}) is seed noise, not
a recipe effect; an out-of-family arm on a much bigger corpus without phase 2 sits between, at
24\%.
Training on $\sim$$59$k non-reasoning rows degrades instruction-following, and part of what looked
like a reasoning regression is a model that still reasons but no longer answers in the requested
form.

The three seed replicates of \S\ref{sec:variance} are themselves three of the six reasoning-only
arms, so the seed spread is represented inside the aggregate rather than hidden from it.
This is \textbf{Qwen only}. For Gpt-OSS the ordering reverses ($62.5$ vs $68.1$) with one run per
condition, so it is unestablished there. Both statements belong in the record: a single A/B is
uninterpretable at this variance, \emph{and} the aggregate is significant.

\textbf{The side experiment's own finding.} Removing the non-reasoning half of the corpus is worth
$+6.9$ pp across 15 runs ($p=0.0008$): the only accuracy effect large and replicated enough to
survive the noise floor, and the only one whose mechanism we can point at. It bears on the main
line in exactly one place: a deployment that never needs a direct mode should train reasoning-only
(\S\ref{sec:recommend}), and that is a deployment choice, not a property of the language results.

\section{Six ways our instruments lied}\label{sec:lied}

\begin{figure*}[t]
\centering
\includegraphics[width=\textwidth]{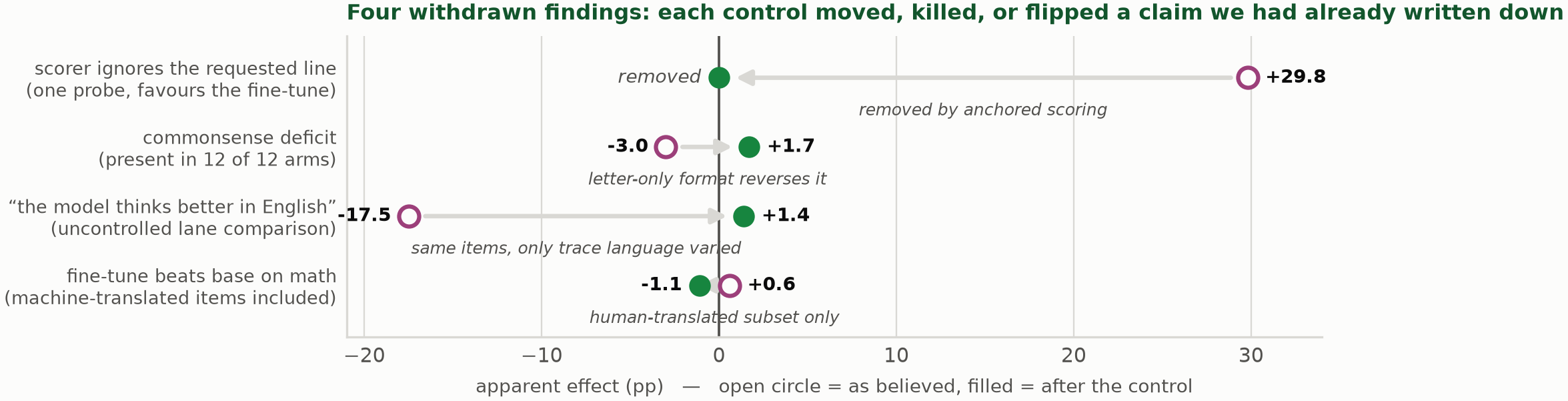}
\caption{The four withdrawn findings of this section, drawn: the effect as first believed (open
circle) against the same quantity after its control (filled). Three flip sign; the un-anchored
scorer artifact ($+29.8$ pp, $18\sigma$, one probe) collapses entirely: its endpoint is drawn
at zero because the prose reports the artifact removed by anchored scoring, not as a measured
residual. Rows are different
instruments and lanes, so magnitudes compare within a row, not across rows; the other pairs'
numbers appear in this section's prose. The section's fifth failure (\S\ref{sec:locale}) is a
scorer defect rather than an effect-size withdrawal, so it has no before/after pair to draw.}
\label{fig:lied}
\end{figure*}

The first four of the following each produced a plausible result that we believed, wrote up, and
withdrew; the fifth was a small correction caught by audit, and the sixth was caught in-flight,
before its wrong numbers reached a table.
Figure~\ref{fig:lied} draws all four before-and-after pairs.

\subsection{A scorer that ignores the requested answer line}

Our prompts request ``\texttt{... write on a new line: The answer is <letter>}''. Our scorer did not
privilege that line; it scanned the whole response and let the last option-mention win. A model that
states its answer and then explains why the others are wrong is therefore scored on its explanation.
The bias is one-directional (verbose arms are penalised, terse arms are not), and on one probe it
produced a $+29.8$ pp ($18\sigma$) artifact favouring the fine-tune. Anchoring extraction to the
requested line removes it. \emph{Any benchmark that requests an answer format must score that format
first and report how often it was absent.}

\subsection{A capability gap that is an artifact of the answer format}

Our headline negative result was a $-3.0$ pp commonsense deficit, present in $12$ of $12$ fine-tuned
arms. Re-asking the same $300$ items with a letter-only prompt reverses it to $+1.7$ pp. The gap
exists only when models reason at length, and the claim as stated was withdrawn. Format must also be
chosen per axis: letter-only is valid for commonsense and invalid for logic, where it drops the base
to $29.6\%$ against a $33\%$ chance baseline.

\subsection{Contamination that a standard check does not find}

Our logic axis was \textbf{38.9\%} contaminated: $175$ of $450$ items shared a 13-gram with the
training pool, while math and commonsense were clean at $0\%$. (The $450$ here is the
\emph{original} logic probe, a hard-subset sampling of the ProofWriter-el pool; the $544$-item
axis in \S\ref{sec:setup} is a different instrument on the same pool; it is the full draw after
removing the $782$ contaminated items. The $38.9\%$ figure is therefore the probe's contamination
rate, and the decontamination cost that left the full pool at $544$.) The cause is structural (the
benchmark and the training slice were drawn from the same ProofWriter pool), and the original build
script verified contamination for the other sources but not this one, because that pool was assumed
eval-only. Removing the items moved individual arms by $0.7$ to $22.2$ pp, i.e.\ \emph{unevenly},
which is what differential memorisation looks like. (The $n$-gram check used throughout is a
surface-match method of the kind \citet{ravaut2024contamination} survey; its known blind spot is
paraphrase, which matters for translated suites, and the Limitations section records that we did not
test beyond it.)

The check also has to run in \emph{both} languages once the training corpus does: the
language-matched corpus carries the English originals of its training questions, so we ran the same
13-gram check between its English half ($21{,}383$--$24{,}509$ rows across the two corpus versions)
and both English evaluation sets (the $1{,}100$-item GSM8K-EN control and the paired English probe).
\textbf{Zero rows collide.}

\subsection{Translating a benchmark with the model family you trained on}

To scale the math axis we machine-translated GSM8K's test split into Greek, keeping $250$
human-translated items (ILSP's MGSM-el set, \S\ref{sec:setup}) as a validity check. The check
failed for exactly one arm (Table~\ref{tab:translation}):

\begin{table}[h]\centering\small
\begin{tabular}{@{}lrrr@{}}
\toprule
arm & human & machine & $\Delta$ \\
\midrule
\textsc{Base} & 95.6 & 90.7 & $-4.9$ ($-3.14\sigma$) \\
\textsc{Subset} & 94.0 & 95.1 & $+1.1$ \\
\textsc{Reasoning} & 93.6 & 95.3 & $+1.7$ \\
\textsc{Two-Phase} & 87.6 & 90.4 & $+2.8$ \\
\bottomrule
\end{tabular}
\caption{The base is $\sim$5 pp worse on our translations; every fine-tune is slightly better.}
\label{tab:translation}
\end{table}

Our training traces and these translations both came from the same commercial model family
(\S\ref{sec:setup}); the translation model is the same one that generated the mathematics
traces. So
the fine-tunes are adapted to its Greek register and the base is not. Including the machine-translated items makes
the fine-tune appear to beat base by $0.6$ pp; on the human-translated subset the base leads by
$1.1$ pp, consistent with an independent benchmark. \emph{If you build a low-resource benchmark by
LLM translation and train on data from the same family, your fine-tune gains an advantage no
contamination check will find: the items are novel, only the register is shared.} A $250$-item
human-translated control costs almost nothing and caught a $3.1\sigma$ artifact.

\subsection{A scorer that cannot read Greek numbers}\label{sec:locale}

The fifth failure is in the scorer that produced every accuracy number in this paper, and we found
it only when auditing the same code for use as a reinforcement-learning reward. Greek, like most of
Europe, writes $17{.}500$ for seventeen-thousand-five-hundred and $3{,}5$ for three-point-five:
the separators are the reverse of the English convention. Our extractor strips commas and calls a
float parse, so a model answering \texttt{114.200} is read as $114.2$ and marked wrong.

On the released Qwen checkpoint's math axis, six answers are written in Greek thousands form and
\textbf{five of them are scored wrong while being right} ($114{,}200$, $17{,}500$, $43{,}200$,
$7{,}300$, $1{,}800$). The mirror case exists and is worse in kind: a decimal comma such as
$3{,}5$ has its separator stripped and becomes $35$, which can score a wrong answer as correct.

The size of the effect on this paper is small: $5$ of $1{,}100$ anchored math rows, $0.45$ pp on
one axis, inside every noise floor we report, and no number in this paper changes. We report it
because the \emph{shape} of the error is the point and generalises past us: the defect is invisible
to an English-language test suite, it is systematic rather than random, and it penalises exactly
the answers written in the target language's own convention. A benchmark built for a
non-English language needs its numeric normalisation tested in that language's conventions, in both
directions, or it will quietly score fluency as failure. \emph{Locale is part of the instrument.}

\subsection{A default that silently changed which lane we measured}\label{sec:lanelie}

The sixth entry was caught while evaluating the RLVR round (\S\ref{sec:rlvr}), by the control this
section keeps recommending: a same-day baseline. The fresh baseline read $1.1\%$ fallback where
the frozen number was $24\%$, too large for noise in either direction. Rescoring the
\emph{original} dump with the current scorer reproduced $24.1\%$ exactly, acquitting the scorer;
the generations themselves differed. The cause was a generation-harness default: without an
explicit flag, the chat template rendered with the reasoning trace \emph{disabled}, so four GPUs
spent four hours generating the direct lane under a filename that said think lane. Nothing
crashed, nothing warned; the numbers were internally consistent and wrong. Every mislabeled dump
was discarded, the flag is now part of the instrument definition next to the locale rule above,
and the incident is why Table~\ref{tab:rlvr} states its \emph{before} column's provenance
explicitly. The general form: \emph{a default that selects which condition you measure is not a
default, it is a hidden factor}, and only a same-day regeneration of a known quantity will
catch it, because every downstream number is plausible.

\keyfinding{\textbf{Finding 4.} Five controls, five withdrawn or corrected findings, and a
sixth failure caught in-flight by the same discipline (\S\ref{sec:lanelie}). Each original
claim was observational or single-condition, and each agreed with what we expected, which is why it
survived review until the control was run. The fifth was found in the scorer itself, and only
because we re-read it for a different purpose (\S\ref{sec:locale}): an instrument can be wrong for
a whole language and still look right on every English test.}

\section{A metric needs a control, not a normalisation}\label{sec:gate}

We tried three definitions of degenerate looping. All were length proxies, with
$|\mathrm{corr}(\text{words},\text{metric})|$ of $0.44$--$0.85$ for a per-token rate and $0.86$--$0.95$
for a fixed-window rate on the arms whose traces are shorter than the window. Normalising by length
does not fix a metric that length causes; the fix is a comparison in which length is held constant by
design, which is what \S\ref{sec:quality} does by conditioning on the item.

We therefore gate every behavioural metric: any candidate correlating $|r| \geq 0.6$ with trace length
is redefined or dropped, never reported with a caveat. The gate removed three of our seven original
dimensions, including a ``the fine-tune loops $29\times$ less'' claim that was pure length artifact.

The measured consequence is worth stating once, because it is asymmetric in a way no length proxy
captures. Applying a repetition penalty suppresses the degenerate loops in both the base and the
fine-tune: at \emph{zero} cost to the base ($-0.4$ pp within the noise floor) and at a $-9.7$ pp
cost to the fine-tune. The base loops \emph{when it has nothing to say}; the fine-tune loops
\emph{as part of how it reasons}: its repeated spans are load-bearing. Removing them does not
clean up its style; it removes part of its argument. The asymmetric price of the same constraint
is the sharpest evidence we have that degeneration means something different in the two models.

\section{Fixing the lock: train on matched language pairs}\label{sec:langmatch}

\begin{figure}[t]
\centering
\includegraphics[width=\columnwidth]{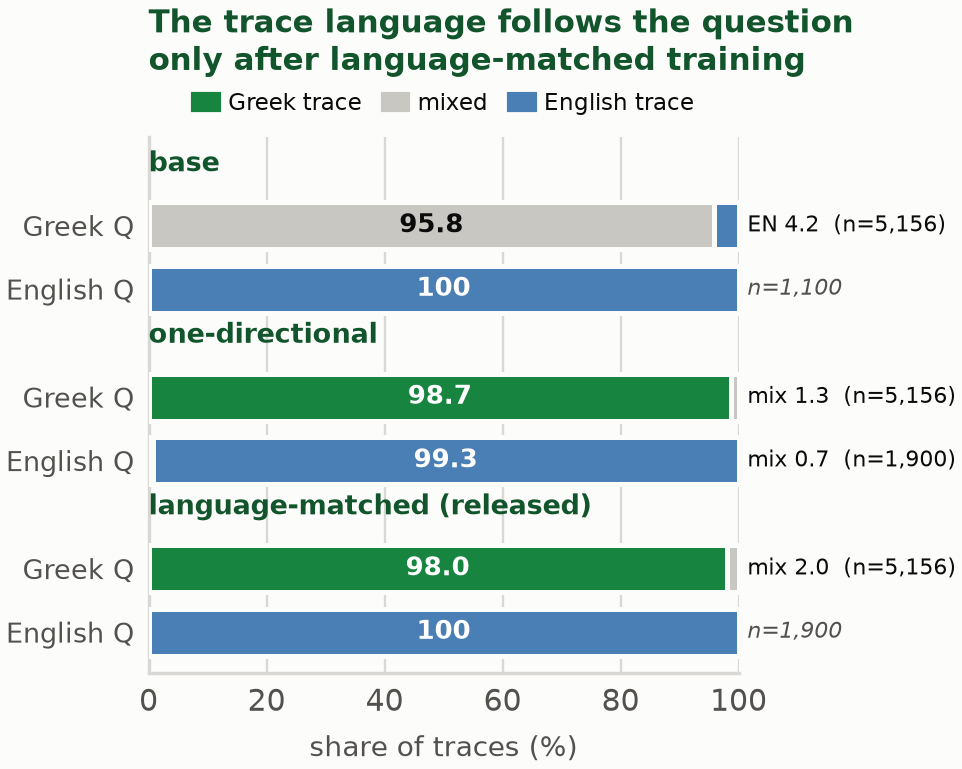}
\caption{Trace-language composition per lane, Qwen line: the share of traces that are Greek
(character ratio $\geq 0.9$), mixed, or English ($\leq 0.1$), on the same instrument as
Table~\ref{fig:families}'s fidelity column. Greek-question rows are the full $5{,}156$-item benchmark;
English-question rows are the $1{,}100$-item control (base) and the $1{,}900$-item probe
(matched). Two honesty notes drawn rather than hidden: the base's Greek-question traces are
\emph{not} mostly pure English: only $4.2\%$ are; the mass is mixed script with English
scaffolding (median ratio $0.33$, \S\ref{sec:quality}); and the one-directional arm's
Greek-question lane decomposes as $98.7$ Greek / $1.3$ mixed / $0.0$ English on the same
instrument ($n{=}5{,}156$, uninstructed); its
English-question lane, generated later on the same $1{,}900$-item probe, measures $99.3\%$
English (Table~\ref{fig:families}). Uninstructed defaults only; the instructed override is
Table~\ref{tab:forget}.}
\label{fig:langmatrix}
\end{figure}

The lock of \S\ref{sec:quality} is not a property of Greek reasoning SFT as such. It is a property
of \emph{training the reasoning language in one direction only}, and it lives in the instruction
channel, not in the default: the one-directional arms still reason in English when the question
is English (Table~\ref{tab:encontrol}), and the locked Qwen arm ignores an explicit instruction
to switch ($0/1{,}000$), a severity that is family-dependent under the same recipe (the
Gpt-OSS one-directional arm stays $95\%$ steerable, \S\ref{sec:families}). A checkpoint
re-trained on \textbf{language-matched pairs} keeps the
question-following default on both lanes for every family we trained, and re-opens the instructed
override on \emph{two of the four} released checkpoints ($44.8$--$62.5\%$ compliance on Qwen and
Gpt-OSS; $0.0\%$ on both Nemotron arms, \S\ref{sec:override}). Figure~\ref{fig:langmatrix} draws the default as
trace-language composition per lane. This
section reports that result self-contained: it is
the answer to the first of three questions we pre-registered (the other two are answered in
\S\ref{sec:plan}, each by an evaluation whose design was fixed before the result), and it changed
the paper's direction.

\textbf{E1: was English reasoning lost, or merely not elicited? Answered in part: not the
capability, and not the recipe alone; the \emph{direction} of training, with a family-dependent
severity we cannot yet explain (the Gpt-OSS arm under the same one-directional recipe stays $95\%$
steerable, \S\ref{sec:families}).} The one-directional fine-tune \emph{looks}
completely locked, and it stays locked when you look harder. Instructed explicitly, on the same
$1{,}000$ items:

\begin{table}[h]\centering\footnotesize
\begin{tabular}{@{}lrr@{}}
\toprule
model, instruction & English traces & Greek traces \\
\midrule
base, ``reason in English'' & \textbf{100.0\%} & 0.0\% \\
base, ``reason in Greek'' & 72.0\% & 0.0\% \\
fine-tune, ``reason in English'' & \textbf{0.0\%} & 98.7\% \\
fine-tune, ``reason in Greek'' & 0.0\% & 98.7\% \\
\addlinespace
matched ft, ``reason in English'' & 44.8\% & 24.8\% \\
matched ft, ``reason in Greek'' & 0.0\% & 98.5\% \\
\bottomrule
\end{tabular}
\caption{The one-directional fine-tune's traces are identical under both instructions, and it
does not matter how the instruction is phrased. Rows classified as neither/mixed scripts ($28\%$
of the base under the Greek instruction) are omitted, so rows need not sum to $100$.
The \emph{matched ft} rows, measured after the language-matched re-training of
\S\ref{sec:langmatch}, show the override partially returning; its mixed share under the English
instruction is $30.4\%$.}
\label{tab:forget}
\end{table}

No prompt we tried recovers the English, so read on this arm alone, the capability looks gone.
It is not gone; the cause is the one-directional recipe, as this section opened. A checkpoint
re-trained on
\textbf{language-matched pairs} (each Greek problem and trace kept beside its English original and
trace, gated so a trace that drifts languages never enters training) reasons in English on
\textbf{100\%} of $1{,}900$ English questions (median Greek-character ratio $0.000$, held across
mathematics, science, medical and general domains), while the English-accuracy cost is $-2.2$ pp
($-2.46\sigma$), half of which is answer-format fallback rather than wrong reasoning. The Greek side
pays more: on the full $5{,}156$-item benchmark the language-matched checkpoint scores $73.7$ against
the base's $80.6$, a $-6.9$ pp raw-lane cost (this is the raw-lane reading; the best-of-$15$ arm
reads $-0.7$ on the $1{,}000$-item probe under answer-format compliance, \S\ref{sec:side}).
That figure sits inside the $7.7$ pp seed floor
measured on this family and decoding (\S\ref{sec:variance}), so we report it as a measured
single-run cost and explicitly not as a property of the recipe. So the
capability was never destroyed by teaching Greek; it was suppressed in the instruction channel (on
the families where it returns at all, \S\ref{sec:override}) by
a recipe that targeted a single output language, while the question-following default survived on
the recipe's sibling arms (Table~\ref{tab:encontrol}). Matching the language of the training pair
to the language of the question keeps that default on every family, and re-opens the instruction
channel on two of the four (\S\ref{sec:override}). Beyond
the default, the explicit override (the capability whose
loss Table~\ref{tab:forget} documents) also returns, partially: re-running the probe that
produced $0/1{,}000$ on the released language-matched checkpoints, an instruction to reason in
English on a Greek question is now obeyed on $44.8\%$ of items (Qwen; $62.5\%$ on Gpt-OSS), with
$24.8\%$ still fully Greek and the rest mixed-script, and the reverse instruction (reason in
Greek on an English question, $n{=}1{,}100$) yields Greek traces on $83.7\%$ / $93.3\%$ (the
reverse lane of the same probe; Table~\ref{tab:forget} carries the Greek-question
rows). Steerability is restored in kind, not in full, and, as \S\ref{sec:override} shows, not on
every family. Two
readings we cannot exclude: partial compliance may be no more than the matched corpus putting
English traces back in-support (a distribution shift rather than restored instruction-following),
and the family ordering ($62.5\%$ over $44.8\%$) reproduces the families' steerability gap under
the one-directional recipe ($95\%$ vs $0\%$, \S\ref{sec:families}), so part of the compliance is
plausibly the family's, not the recipe's.
The word the paper is allowed to use is therefore
\textbf{language-matching}, not language-locking: the reasoning language follows the question.

\subsection{The instruction channel re-opens on two families, not four}\label{sec:override}

Running the identical probe, scorer and thresholds on the two Nemotron language-matched releases
completes the picture, and it is not the one the two-family result suggested:

\begin{table}[h]\centering\footnotesize
\setlength{\tabcolsep}{4pt}
\begin{tabular}{@{}lrr@{}}
\toprule
released checkpoint & ``in English''$\uparrow$ & ``in Greek''$\uparrow$ \\
 & (Greek Q) & (English Q) \\
\midrule
Sophea-Qwen3.6-v1 & 44.8\% & 83.7\% \\
Sophea-Qwen3.6-v1.1 (RLVR) & 53.9\% & 85.7\% \\
Sophea-OSS-v1 & \textbf{62.5\%} & 93.3\% \\
Sophea-Nemo-3.5-Lightning-v1 & \textbf{0.0\%} & 92.7\% \\
Sophea-Nemo-3-Nano-v1 & \textbf{0.0\%} & 87.5\% \\
\midrule
\emph{one-directional (reference)} & \emph{0.0\%} & \emph{91.4\%} \\
\bottomrule
\end{tabular}
\caption{Instructed-override compliance on the five released checkpoints: the share of traces in
the instructed language when the instruction contradicts the question ($n{=}1{,}000$ Greek-question
and $n{=}1{,}100$ English-question items; same scorer as Table~\ref{tab:forget}). Every release
obeys ``switch to Greek''; only the Qwen line and Gpt-OSS obey ``switch to English'', and the
RLVR refinement widens Qwen's opening ($44.8\to53.9\%$, \S\ref{sec:rlvrresults}). Two instrument notes,
both checked: the Nemotron dumps do not separate trace from answer, so the mandatory Greek answer
line falls inside the scored text: re-scoring with that line stripped leaves both Nemotron arms
at $0.0\%$ (the Qwen control moves $44.8 \to 49.5\%$, the expected direction); and the Nemotron
arms return $5$--$9\%$ of rows too short to score against Qwen's $0\%$, consistent with the low
reasoning ceiling reported in \S\ref{sec:families}.}
\label{tab:override}
\end{table}

So the instruction channel is \emph{not} a property the language-matched recipe restores. It
restores the question-following default on every family we trained (Table~\ref{fig:families}), and
re-opens the override on two of four checkpoints. The asymmetry is uniform in one direction and
family-dependent in the other: every release complies with an instruction to reason in Greek
($83.7$--$93.3\%$), none complies fully with an instruction to reason in English, and the two
Nemotron arms do not comply at all. The split tracks neither
architecture (the two Nemotron generations differ from each other in reasoning ceiling but agree
here) nor recipe (identical). What we can now say (from the pre-registered RLVR round of
\S\ref{sec:rlvr}) is that the channel \emph{responds to a verifiable reward}: an
override-obedience term moves compliance $+9.1$\,pp on this family with a flat random-reward
control and no cost on any held direction, though short of the pre-registered trainability bar
(\S\ref{sec:rlvrresults}). Why the channel sits open on
two families and shut on two others after identical SFT remains the open question; whether it
moves under a reward at all no longer is.

The practical consequence is the part a deployment can act on. Trace-language \emph{steerability}
is not a property this recipe confers, and it is not predictable from the axes a model card
publishes: the four checkpoints here share a corpus, a recipe, an adapter configuration and a
serving budget, agree on every default-behaviour metric we report, and still split $62.5/44.8/0/0$
on this one. \textbf{A deployment that needs the trace language to be steerable at inference (a
bilingual support desk, a reviewer-facing audit mode, any product where the operator overrides the
user's language) must test this axis per checkpoint, and must not infer it from the recipe, the
family, or the question-following behaviour that this paper otherwise shows generalising.} The
probe costs one instructed pass over a benchmark that already exists ($n{=}1{,}000$ per direction
here), which puts it in the same class as the seed control of \S\ref{sec:variance}: cheap, and it
changes a claim we would otherwise have made. We add it to the supported column of
Table~\ref{tab:recommend} on those terms.

The honest reading of Table~\ref{tab:forget} is therefore not ``the model forgot English'' but ``a
model that targets a single output language learns to ignore explicit language
instructions.'' Note the base is partially locked too (told to reason in Greek it still
produces English $72\%$ of the time), so asymmetric locking is not unique to fine-tuning; what is
unique is how firm ours became.

The finding is not Qwen-specific, and it is not specific to the language-matched checkpoint. Running
the same $1{,}100$-item English control (the GSM8K originals of the Greek math axis,
anchored-extraction scoring, \S\ref{sec:lied}) on the release candidates of all three families, every fine-tuned
arm still reasons in English on $100\%$ of English questions (median Greek-character ratio $0.00$),
at $+0.6$ to $+2.2$ pp against base for the reasoning-recipe arms; the matched-recipe rows stand
at $+0.4$ (Qwen) and $-3.6$ (Gpt-OSS) against their own bases, with the Gpt-OSS English retention
since re-measured at $+1.0$ macro against
base (Table~\ref{tab:forgetting}):

\begin{table}[h]\centering\small
\begin{tabular}{@{}lcc@{}}
\toprule
arm ($n{=}1{,}100$) & accuracy & reasons in EN \\
\midrule
Qwen base / ft    & 94.0 / 96.2 & 100\% / 100\% \\
Gpt-OSS base / ft & 95.5 / 96.1 & 100\% / 100\% \\
NemotronH base / ft & 92.7 / 93.7 & 100\% / 100\% \\
\midrule
Qwen language-matched & 94.4 & 100\% \\
Gpt-OSS language-matched & 91.9 & 100\% \\
\bottomrule
\end{tabular}
\caption{English questions get English traces from every arm measured. The fine-tunes in the
first three rows are the dual-mode / reasoning-recipe checkpoints (\S\ref{sec:recipes}), kept
deliberately: they are the recipe-class evidence that one-directional training does not lock
English \emph{questions}. The last two rows are the language-matched checkpoints: the released
Qwen checkpoint, re-run on this exact control ($94.4$, $+0.4$ over its base, $100\%$ English
traces, consistent with its $1{,}900$-item probe above), and the Gpt-OSS language-matched arm
evaluated after merging (the pre-repair arm; the released, format-repaired checkpoint's control
has not been generatively re-run, Table~\ref{fig:families}). Together the rows confirm the
language-matching result is not Qwen-specific; none of this retests the hard one-directional
lock of Table~\ref{tab:forget}, which remains a property of training in a single output direction.}
\label{tab:encontrol}
\end{table}

A question the printed numbers force: on our printed axes the one-directional fine-tunes match or
beat the matched checkpoint: higher Greek-trace fidelity ($98.7$ vs $97.98$), the same $100\%$
English-on-English default (theirs measured on sibling arms of the same recipe,
Table~\ref{tab:encontrol}; no single one-directional arm carries all three numbers at once), and
no $-6.9$ print. What they lack is the instruction channel: the override probe reads $0/1{,}000$
on the one-directional arm and $44.8\%$ English compliance on the matched checkpoint ($62.5\%$ on
the Gpt-OSS release; Table~\ref{tab:forget}). We release the matched checkpoints as the headline
because default behaviour and instructability are different axes: the matched recipe is the only
one measured to restore the default on every family, and the only one under which the override
returns at all, though on two of four checkpoints, not four (Table~\ref{tab:override}). The
release rationale therefore stands as ``defaults kept everywhere, overrides re-opened on two
families'': measured, no longer a hedge, and no longer stated as a property of the recipe alone.

\section{Two pre-registered questions, and how they closed}\label{sec:plan}

The language-matching result of \S\ref{sec:langmatch} answered the first of the three questions we
pre-registered. The other two closed since, each by a cheap evaluation whose design was fixed
before the result existed; we state each design first and its answer second, so the results read
as a test rather than a story fitted afterwards.

\textbf{E2: Did Greek reasoning SFT damage general ability, in either language? Answered: no on
the earlier recipe, with one repairable exception on the language-matched recipe.} (One prior
frames the null: LoRA is documented to forget less than full fine-tuning at matched target gain
\citep{biderman2024lora}, so flat macros are the expected case, not a surprise; reported
anyway, because a 35B MoE at rank 32 on a translated suite is a different setting from theirs.)
The corpus carries $5{,}211$ English rows precisely as replay, and the
risk on the Greek side is subtler: a model taught to reason at length may lose the short, direct
competence that general benchmarks measure. The instrument is fixed by the data, not chosen for
convenience. Every arm's non-reasoning half \emph{is} the \textbf{Sophea-Titan-1} mix
(\S\ref{sec:setup}), so the retention question has to be asked on the suite that model was released
against, or the numbers compare to nothing: nine frozen Greek benchmarks (GreekMMLU, MMLU-el,
HellaSwag-el, ARC-Easy/Challenge-el, Belebele-el, Winogrande-el, TruthfulQA-el, Medical-MCQA-el) and
five English-retention benchmarks (MMLU, HellaSwag, ARC-Easy/Challenge, Winogrande), all scored by
log-likelihood with no generation involved. Two macros, deliberately not averaged into one: the
Greek macro answers whether reasoning SFT cost general Greek ability, the English macro whether it
cost English.

\begin{table}[h]\centering\footnotesize
\setlength{\tabcolsep}{3.5pt}
\resizebox{\columnwidth}{!}{%
\begin{tabular}{@{}lccc@{}}
\toprule
Greek macro & base & language-matched & \textbf{+ format repair} \\
\midrule
Gpt-OSS & $61.2$ & $53.9$ & $\mathbf{58.0}$ \\
\midrule
\multicolumn{4}{@{}l@{}}{\emph{macro deltas vs base (points):}}\\
Sophea-Qwen3.6-v1 & \multicolumn{3}{c}{$-0.01$ Greek / $+0.08$ English} \\
Sophea-OSS-v1 & \multicolumn{3}{c}{$-3.2$ Greek / $+1.0$ English$^{\dagger}$} \\
Sophea-Nemo-3-Nano-v1 & \multicolumn{3}{c}{$+3.8$ Greek / $-2.0$ English} \\
Sophea-Nemo-3.5-Lightning-v1 & \multicolumn{3}{c}{$+1.7$ Greek / $-1.1$ English} \\
\bottomrule
\end{tabular}}
\caption{Forgetting on the Titan-1 suite. Top: Gpt-OSS on the nine Greek non-reasoning benchmarks,
base against the language-matched fine-tune, before and after the format-repair dose: the
language-matched fine-tune initially costs $-7.3$ points, and the repair dose recovers $4.1$ of
them, so the residual is $-3.2$ not zero. Below: the macro deltas this paper reports per family, fine-tuned arm vs its own
base; \textbf{the format-repaired Gpt-OSS is the fine-tuned arm we stand behind}. Qwen is flat
in both languages; Nano \emph{gains} Greek (+3.8) against the family's low 0.49 Nano base (mostly
catch-up, not forgetting); Lightning is read against its \emph{own} base, measured separately
(Greek macro 57.5, well above Nano's 48.6), and gains +1.7; an earlier +10.6 reading against
the shared Nano base is superseded by that measurement. Per benchmark
against binomial standard error; a macro delta is a direction, not a test.
$^{\dagger}$On English the pre-repair language-matched Gpt-OSS lost $-7.7$ points ($75.7 \to 68.0$);
re-measured on the repaired release, the English macro is $76.7$, $+1.0$ \emph{above} base: the
repair dose recovered both lanes, so the pre-repair loss was format behaviour throughout.
Macro deltas are computed on unrounded macros; Table~\ref{tab:forgetbench} prints one decimal,
so its Qwen columns round to $75.7{\to}75.6$ Greek / $85.1{\to}85.2$ English.}
\label{tab:forgetting}
\end{table}

\begin{table*}[t]\centering\footnotesize
\setlength{\tabcolsep}{4.5pt}
\begin{tabular}{@{}lrrrrrrrr@{}}
\toprule
 & \multicolumn{2}{c}{Qwen} & \multicolumn{2}{c}{Gpt-OSS} & \multicolumn{4}{c}{Nemotron} \\
\cmidrule(lr){2-3}\cmidrule(lr){4-5}\cmidrule(lr){6-9}
benchmark & base & ft & base & ft & Nano base & Nano ft & Ltng.\ base & Ltng.\ ft \\
\midrule
\multicolumn{9}{@{}l}{\emph{Greek (nine benchmarks, the Greek macro of Table~\ref{tab:forgetting}):}}\\
ARC-Challenge-el & 91.8 & 92.0 & 77.1 & 68.6 & 54.2 & 59.0 & 68.7 & 72.8 \\
ARC-Easy-el & 96.9 & 97.1 & 85.4 & 79.9 & 62.3 & 70.1 & 75.8 & 81.4 \\
Belebele-el & 92.7 & 93.7 & 86.0 & 81.3 & 67.0 & 78.0 & 81.3 & 83.3 \\
GreekMMLU & 83.8 & 83.0 & 66.3 & 61.3 & 58.8 & 53.3 & 64.2 & 62.2 \\
HellaSwag-el & 57.6 & 59.4 & 42.9 & 52.1 & 39.2 & 47.6 & 44.5 & 51.2 \\
Medical-MCQA-el & 81.2 & 80.3 & 42.1 & 38.7 & 27.8 & 31.2 & 41.7 & 38.2 \\
TruthfulQA-el & 39.1 & 36.5 & 39.7 & 35.5 & 31.7 & 33.2 & 34.2 & 33.7 \\
Winogrande-el & 61.0 & 62.5 & 55.4 & 59.6 & 53.3 & 58.3 & 55.8 & 60.1 \\
MMLU-el & 76.9 & 76.2 & 55.5 & 45.4 & 43.2 & 40.8 & 51.3 & 49.6 \\
\midrule
\emph{Greek macro} & \emph{75.7} & \emph{75.6} & \emph{61.2} & \emph{58.0} & \emph{48.6} & \emph{52.4} & \emph{57.5} & \emph{59.2} \\
\addlinespace
\multicolumn{9}{@{}l}{\emph{English retention (five benchmarks):}}\\
ARC-Challenge & 96.1 & 95.7 & 90.0 & 86.8 & 88.3 & 82.2 & 92.0 & 89.1 \\
ARC-Easy & 99.1 & 99.1 & 96.0 & 95.7 & 96.2 & 93.4 & 97.8 & 96.4 \\
HellaSwag & 73.2 & 74.2 & 57.0 & 68.4 & 67.9 & 73.1 & 72.7 & 75.9 \\
Winogrande & 73.2 & 74.2 & 64.6 & 66.7 & 69.9 & 72.9 & 72.0 & 73.3 \\
MMLU & 83.9 & 82.6 & 70.8 & 65.8 & 69.9 & 60.5 & 75.8 & 69.7 \\
\midrule
\emph{English macro} & \emph{85.1} & \emph{85.2} & \emph{75.7} & \emph{76.7} & \emph{78.4} & \emph{76.4} & \emph{82.0} & \emph{80.9} \\
\bottomrule
\end{tabular}
\caption{Every benchmark behind the macros of Table~\ref{tab:forgetting} (Titan-1 suite,
non-reasoning mode; accuracy \%). Qwen \emph{ft} is Sophea-Qwen3.6-v1 (language-matched); Gpt-OSS
\emph{ft} is Sophea-OSS-v1, the format-repaired release (its pre-repair intermediate is in
Table~\ref{tab:forgetting}); Nano ft is Sophea-Nemo-3-Nano-v1, the language-matched Nano release;
Lightning ft is Sophea-Nemo-3.5-Lightning-v1 (language-matched); the two
Nemotron generations share the same 31.6B / 6-of-128 routing profile (Table~\ref{tab:models})
and each is read against its own base column. (The earlier one-directional recipe arm is reported
only in \S\ref{sec:families}'s recipe history, not as a release.) The per-benchmark view localises the Gpt-OSS story: before the repair the loss
concentrated on the instruction-format-heavy reading benchmarks (Belebele-el $-17.7$,
ARC-Challenge $-13.1$ el / $-19.1$ en, MMLU-en $-16.0$) while HellaSwag and Winogrande
\emph{gained} in both languages, and the repair dose recovers most where the loss was format
(Belebele-el regains $+13.0$ of the $17.7$). English retention on the released Gpt-OSS
checkpoint recovers to $+1.0$ above its base; Lightning retains English within $1.1$ points of
its own base.}
\label{tab:forgetbench}
\end{table*}

\begin{figure}[t]
\centering
\includegraphics[width=\columnwidth]{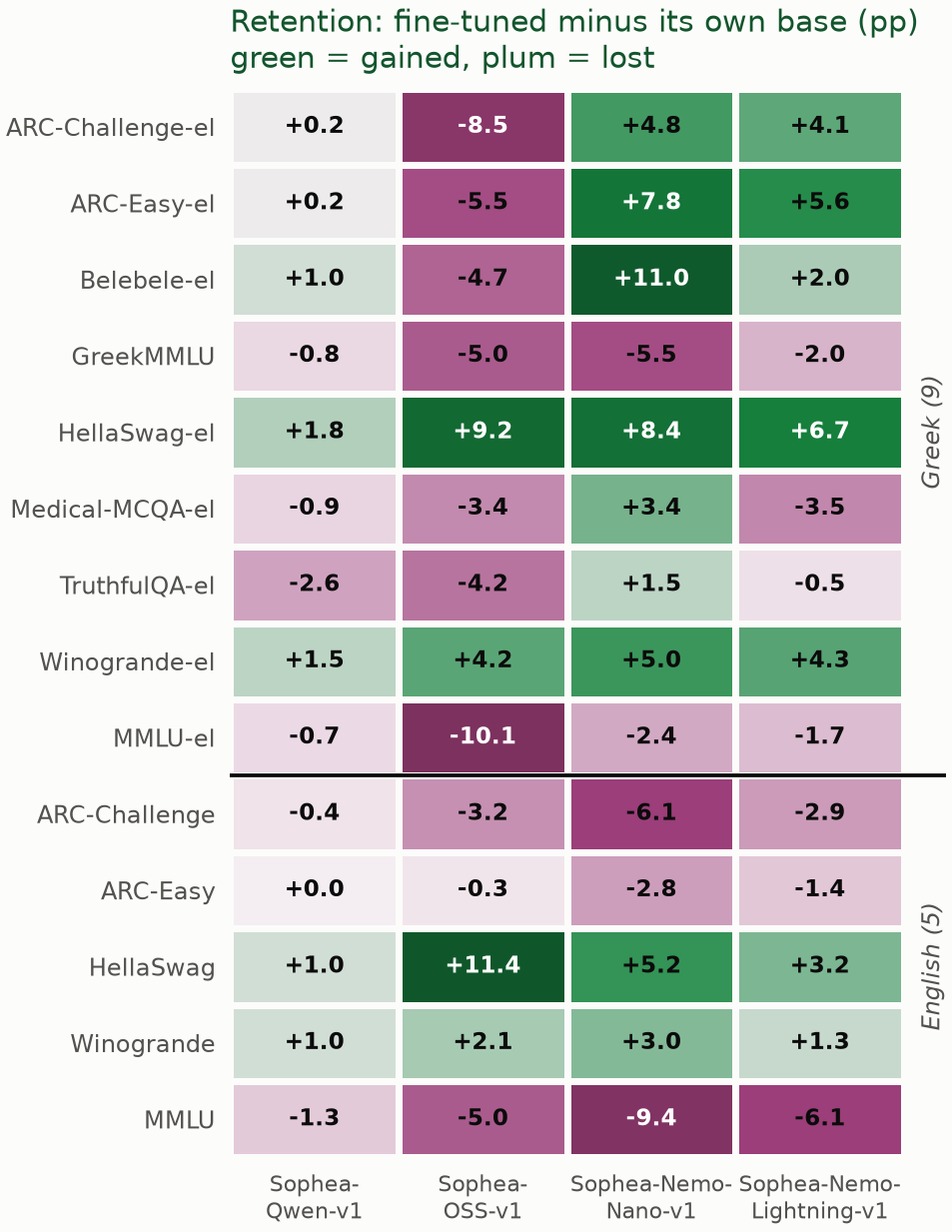}
\caption{Every per-benchmark retention delta behind Table~\ref{tab:forgetbench}: fine-tuned
release minus \emph{its own} base (Lightning against the Lightning base; Sophea-OSS-v1 is the
format-repaired release), Titan-1 suite, non-reasoning mode. Per-benchmark binomial sigma
applies and a delta is a direction, not a test (Table~\ref{tab:forgetting}); the value is the
pattern: the Qwen column is near-white in both languages, the Sophea-OSS-v1 residual
concentrates on the knowledge-heavy benchmarks (MMLU-el $-10.1$) while HellaSwag gains in both
languages on every family, and MMLU is the largest English loss for all three non-Qwen
releases.}
\label{fig:forgetdelta}
\end{figure}

Table~\ref{tab:forgetting} reports the release reasoning models this paper stands behind
(Table~\ref{tab:forgetbench} opens every macro into its per-benchmark readings;
Figure~\ref{fig:forgetdelta} draws the same deltas as one map). A reader
comparing the \emph{base} columns against the numbers the labs publish for the same checkpoints
should expect small offsets, on the order of $\pm 1$--$2$ points: published tables come from
different evaluation stacks: serving engine, prompt template, few-shot count, sampling
parameters (temperature, top-$p$) where generation is involved, and answer-extraction rules.
None of those differences is a property of the model. Every arm in these tables is scored by
the same harness under the same settings, so the columns are internally comparable; it is the
base-to-fine-tune \emph{delta}, not the absolute level, that this section's claims rest on.
Sophea-OSS-v1 is the column that matters. Sophea-Qwen3.6-v1 is flat to a tenth of a point in both languages.
On Sophea-OSS-v1 the language-matched fine-tune is not uniformly free: it costs $-7.3$ points Greek
and $-7.7$ English (macro $75.7 \to 68.0$) \emph{in both reasoning and direct modes}, with a $70\%$ answer-fallback rate.
Adding a format-repair dose (the identical mix plus a $2\%$ dose of trace-less anchored
answer-closing twins) recovers $4.1$ of the Greek points and collapses fallback from $70\%$ to
$26\%$ on the full benchmark (per-domain in Table~\ref{tab:domain}),
so the Sophea-OSS-v1 deficit is format-closing behaviour, fixable by data, not a capability the
fine-tune destroys. We therefore release \textbf{the
format-repaired Gpt-OSS as Sophea-OSS-v1}; the cost that survives is $-3.2$ Greek. English
retention on the released checkpoint, re-measured on the same five benchmarks, reads $76.7$
($+1.0$ \emph{above} its base), so the pre-repair $-7.7$ was the same format-closing behaviour on
the English lane, and the Greek-side repair dose recovered both. The replay did its job on Sophea-Qwen3.6-v1 and
Sophea-Nemo-3-Nano-v1, and the language lock in E1 is a preference, not damage; the
single family-specific cost is real but repairable.

\textbf{Register and grammar, probed directly.} Two small judge-scored probes ask what the NLU
macros cannot: does the fine-tune still switch address register on request (the formal
plural-of-politeness vs the informal singular; $46$ items spanning single-turn, rewrite and
mid-conversation switches), and is its free-form Greek grammatical (gender/number/case agreement, aspect,
voice/mood, clitic placement; $58$ items)? Register control survives fine-tuning in every family
and improves where the base was weakest (the two Nemotron bases go $40/46$ and $37/46$ to $44/45$
and $43/46$; Qwen dips $45 \to 42$, Gpt-OSS is flat at $39/46$). Grammaticality roughly
\emph{doubles} on both Nemotron fine-tunes ($13/58 \to 27$ and $12/58 \to 29$, driven by the
agreement axis), is flat on Qwen ($41 \to 42$), and improves on Gpt-OSS ($32 \to 40$). The
Gpt-OSS cell also demonstrates why these probes need adversarial reading: its first-pass score
was $23/58$, a spurious regression caused by the same channel-marker behaviour documented above:
the fine-tune prefixes its text with a bare channel marker, the judge read the marker as
ungrammatical Greek, and stripping it flipped the verdict from ``worst fine-tune'' to ``above
base''. At $n{=}46$/$58$ these are counts, not
rates, and the scorer is itself an LLM judge (temperature $0$), so we report them as directional
evidence only: the behavioural shift this paper documents costs neither politeness control nor
grammatical Greek in any family.

\textbf{E3: Does the non-reasoning mode still work? Answered: yes; dropping the direct half
costs nothing measurable there.} The direct half exists to preserve that mode, its removal is our
largest measured accuracy effect, and every other number in this paper comes from reasoning
benchmarks, so the design was fixed in advance: run the Greek NLU benchmark (instrument (v),
\S\ref{sec:setup}) in direct mode, no reasoning requested, across \textsc{Base},
\textsc{Reasoning} (which never saw the direct half) and \textsc{Two-Phase} (which did). This was
the one place where dropping phase 2 should \emph{cost} something. It does not: on strict scoring
Qwen reads $0.680$ base, $0.683$ reasoning-only, $0.696$ two-phase (a gap inside the seed floor),
and on Gpt-OSS the reasoning-only arm sits at base level ($-1.6$ pp) while \emph{two-phase} is the
only arm that degrades ($-9.0$ pp strict, and the one format break: its direct-mode outputs carry
English reasoning prose the strict parser rejects). The release checkpoints were then re-run on
the Titan-1 suite in the same mode: Qwen flat ($-0.1$), Nemotron-3.5 $+1.7$ against its own base
(an interim $+10.6$ reading, taken against the shared Nano base before the Lightning base was
measured, is superseded), and pre-repair Gpt-OSS $-7.3$: the same $-7.3$ as its think mode, so that family's deficit
is mode-independent, exactly what the format-repair reading of Table~\ref{tab:forgetting}
predicts. Two by-products belong in the record. The \texttt{<think>}-carrier families have a clean
off switch: zero unrequested traces in $9{,}751$ direct-mode generations. The Gpt-OSS \emph{base}
has no off switch at all (it opens its analysis channel on $100\%$ of rows even with reasoning
effort set to none), and neither of its fine-tunes installed a switch the base never had.

\keyfinding{\textbf{Why these designs.} Each is decidable by evaluation alone, needs no retraining,
and would change a sentence this paper commits to. (E1, the question that recast the goal from
locking to matching, is answered in \S\ref{sec:langmatch}.) E2 closed the forgetting question:
general ability was flat or gained on Sophea-Qwen3.6-v1 and Sophea-Nemo-3-Nano-v1, and the one release
that lost ground (Sophea-OSS-v1) was shown to lose it to format-closing behaviour and to recover
under a format-repair dose, not to a destroyed capability. E3 closed the direct-mode question the
same way: the design was fixed while either outcome was still possible, and the answer (dropping
the direct half costs nothing measurable on the non-reasoning mode) landed on the side that
strengthens \S\ref{sec:recommend} rather than the side we could not have walked back.}

\section{Per domain: indistinguishable on accuracy, not on cost}\label{sec:domain}

\begin{figure*}[t]
\centering
\includegraphics[width=\textwidth]{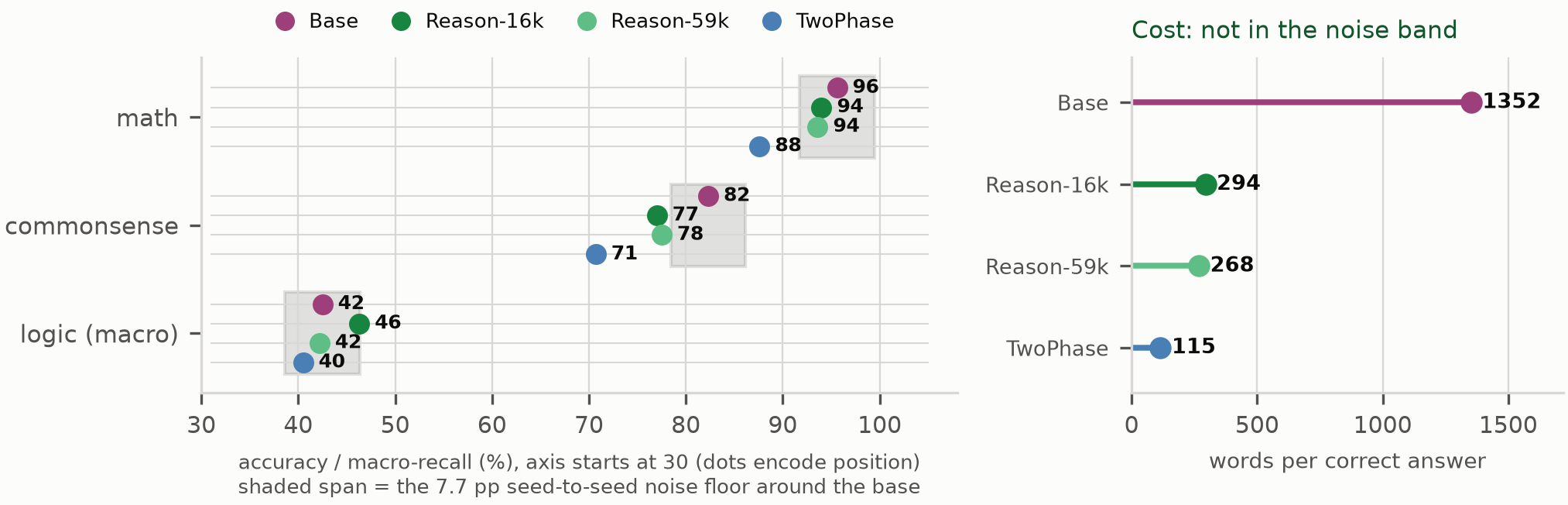}
\caption{Left: accuracy per axis on the 5{,}156-item benchmark for the \emph{Qwen recipe arms}
(the cross-family releases are Table~\ref{tab:domain}), with the seed-to-seed noise floor shaded
around the Qwen base. Right: the same arms' cost in \emph{words} per correct answer (token-denominated
cost is Figure~\ref{fig:tokens}). Math uses the 250 human-translated items only
(\S\ref{sec:lied}); logic is macro-recall (\S\ref{sec:metrics}).}
\label{fig:domain}
\end{figure*}

Figure~\ref{fig:domain} is deliberately not a leaderboard. At a 7.7 pp seed range, naming a
per-domain winner selects the top of a noisy draw, and a rerun would name someone else; the shaded
band makes that visible rather than leaving it to a footnote. Read that way it says two things.

\textbf{On accuracy, no per-domain winner is readable against the floor.} Math 87.6--95.6,
commonsense 70.7--82.3, logic 40.5--46.3: spreads of 8.0, 11.6 and 5.8 pp against a 7.7 pp seed
range, two of them slightly wider than the range itself. Choosing a model per domain on these
numbers is not supported by this evidence.

\textbf{On cost, the arms are not close.} 1{,}352 \emph{words} per correct answer for the base
against 268--294 for the reasoning-only fine-tunes: a 4.6--5.0$\times$ difference, far outside
anything seeds do, though in tokenizer tokens the gap narrows or reverses by family
(\S\ref{sec:quality}). If a deployment must pick one Greek reasoning model, the defensible basis is
cost and trace language, not per-axis accuracy.

\section{What each family taught us}\label{sec:families}

The three models behaved differently in ways that no accuracy table records, and that anyone
adapting them will meet. We report these because they cost us real time and are absent from the
model cards. Library versions are peft 0.19.1 / transformers 5.12.1 throughout; items marked
\emph{[env]} are version-specific rather than properties of the model.

\subsection{Qwen3.6-35B-A3B: fused experts, and a base that hides behind its token budget}

\emph{[env]} Its experts are \textbf{stacked parameters}, so LoRA attaches via
\texttt{target\_parameters} (28 tensors). That adapter \textbf{cannot be loaded back} by
\texttt{PeftModel.from\_pretrained}: the call raises inside a weight converter, and the only route
to evaluation is merging the adapter into dense weights first. We lost an entire evaluation round to
this before noticing that the base arms had run and the fine-tuned arms had not.

\textbf{Its base is truncation-crippled in think mode.} At $1{,}536$ tokens, $670$ of $1{,}000$ rows
never reach a final answer and it scores $20.8$; at $4{,}096$ it scores $77.2$ (the
$1{,}000$-item probe's three-axis mean, as in \S\ref{sec:intro}), above its own direct
mode. Any comparison against this base at a standard budget measures the budget. Its traces run
$4{,}000$--$5{,}400$ characters where the fine-tune's run $150$ words.

\textbf{It is the most seed-sensitive of the three} (sd $4.4$ pp, range $7.7$), and its
\emph{one-directional} fine-tune is
completely \textbf{language-locked}: told to reason in English it produces Greek on $1000/1000$
items (the released language-matched checkpoint complies on $44.8\%$, Table~\ref{tab:override};
$53.9\%$ after the RLVR round, Table~\ref{tab:rlvr}). The base is partially locked the other way:
told to reason in Greek it still emits English $72\%$ of the time.

\subsection{Gpt-OSS-20B: a different trace carrier, and it never stops reasoning}

\textbf{It does not use \texttt{<think>}.} Reasoning arrives in \emph{harmony channels}
(\texttt{analysis} / \texttt{assistantfinal}), so a scorer written for one carrier silently
mis-reads the other.

\textbf{It emits an analysis channel even when reasoning is switched off}: $1000/1000$ rows at
\texttt{reasoning\_\allowbreak effort=none}. We gated the channel split on having \emph{asked} for reasoning
rather than on the markers being present, so every direct-mode number was computed over trace and
answer concatenated, and $95$ logic rows scored as unparseable. Split whenever the markers appear.

\emph{[env]} A \texttt{detailed thinking off} system message does not disable reasoning; it returns
\texttt{content=null} with the text in \texttt{reasoning}. Use
\texttt{chat\_\allowbreak template\_\allowbreak kwargs}. It also rejects \texttt{sdpa} and needs \texttt{eager} attention.

\textbf{It is the counterexample twice over.} Dropping the non-reasoning half of the corpus helps
Qwen by $+6.9$ pp and \emph{hurts} Gpt-OSS by $5.6$ ($62.5$ vs $68.1$, one run per condition,
\S\ref{sec:side}); and its fine-tune stays \textbf{steerable},
complying with an English-reasoning instruction $95\%$ of the time where Qwen's complies $0\%$
(the released language-matched arms narrow the gap but keep the ordering: $62.5\%$ vs $44.8\%$,
\S\ref{sec:langmatch}). Same
recipe, same corpus, opposite behaviour. We have no mechanism for either difference, though
\S\ref{sec:rlvrresults} shows the override quantity itself responds to a verifiable reward.

\subsection{NemotronH-30B-A3B: a hybrid that fights every default}

\begin{sloppypar}
\textbf{Its experts are individually materialised as \texttt{nn.Linear}},
so PEFT attaches one adapter \emph{per expert}: \textbf{4{,}188 adapter tensors} against 28 for
Qwen, at \emph{fewer} trainable parameters (325M vs 688M). Step time then goes on kernel launches
and gradient all-reduces rather than FLOPs; it trains $3.7$--$7.6\times$ slower than the others.
Reducing adapter count $7\times$ bought only $23\%$, so this is not the whole story; we did not
isolate the rest.
\end{sloppypar}

\begin{sloppypar}
\emph{[env]} Three defaults have to change, none of which announces itself.
\texttt{ddp\_find\_\allowbreak unused\_\allowbreak parameters} must be \textbf{True} (routing leaves experts
without gradient, and DDP aborts mid-run); \texttt{model\_accepts\_\allowbreak loss\_\allowbreak kwargs} must be
\textbf{False} (NemotronH ignores \texttt{num\_items\_\allowbreak in\_\allowbreak batch}, so loss and gradients otherwise
scale with the accumulation count); and evaluation needs \texttt{trust\_remote\_\allowbreak code=False},
because the shipped
modeling file \emph{trains} correctly but its \texttt{prepare\_\allowbreak inputs\_\allowbreak for\_\allowbreak generation} indexes a
\texttt{cache\_position} that \texttt{generate()} passes as \texttt{None}. That last one fails only
at inference, so five hours of training completed before it surfaced.
\end{sloppypar}

\emph{[env]} \texttt{mamba-ssm} must be installed or 23 of its 52 layers fall back to a pure-PyTorch
scan; transformers logs this on every run. Installing it changed step time by $0\%$;
worth recording, because the obvious diagnosis was wrong.

\textbf{And the claim replicates on it.} This is the third, non-transformer architecture, and the
headline result holds: its base reasons in Greek on $0\%$ of the Greek benchmark (median ratio
$0.23$), its fine-tune on $98\%$, at an accuracy cost of $-0.6$ pp, with the generation-cap
truncation rate falling $10.8 \to 2.6\%$ and measure loops in its traces dropping from $15.0$ to
$0.0$ per thousand rows (same $5{,}156$-item benchmark, full lanes on both arms). Whatever made the
reasoning language movable on two transformer MoEs is not a transformer property; on this
evidence it travels with a strong base plus a language-directed corpus.

\keyfinding{\textbf{The pattern.} Every family needed a different fix, none of the fixes was
announced by an error at the point of the mistake, and three of them produced \emph{plausible
numbers} rather than crashes. A pipeline that runs to completion on a new model family is not
evidence that it ran correctly.}

\section{What we can and cannot recommend}\label{sec:recommend}

Table~\ref{tab:recommend} states every configuration recommendation this evidence supports, and
the ones it cannot; the basis column is the point, because a configuration table without it would
be the overclaim this paper argues against.

\begin{table*}[t]\centering\footnotesize
\setlength{\tabcolsep}{4pt}
\begin{tabular}{@{}p{8.6cm}p{7.2cm}@{}}
\toprule
\textbf{Supported by our evidence} & \textbf{basis} \\
\midrule
Drop the hybrid phase 2; train reasoning-only & +6.9 pp over 15 runs, $p=0.0008$ (\S\ref{sec:side}); and the direct mode it sacrifices costs nothing measurable on direct-mode NLU (E3, \S\ref{sec:plan}) \\ \rowsep
Generate exploratory traces, not explanations & structure $0.27 \to 0.68$ on the same questions \\ \rowsep
Forbid formulaic openings in the generator & 24/24 identical openings without it \\ \rowsep
Report \texttt{fallback\%} beside accuracy & separates reasoning from format compliance \\ \rowsep
Run the seed control before any ablation & 7.7 pp (\S\ref{sec:variance}) \\ \rowsep
Probe trace-language steerability per checkpoint & same recipe, four checkpoints, $62.5$/$44.8$/$0$/$0$ (\S\ref{sec:override}) \\ \rowsep
Keep a human-translated benchmark subset & caught a $3.1\sigma$ artifact (\S\ref{sec:lied}) \\ \rowsep
Test numeric normalisation in the target language's own convention & $5$ of $6$ Greek-thousands answers scored wrong while right (\S\ref{sec:locale}) \\ \rowsep
Gate metrics on length correlation & removed 3 of our 7 \\ \rowsep
Fix format and leak defects with verifiable-reward RL, not more SFT data & fallback $24\!\to\!2.5\%$, leak $3.5\!\to\!0\%$, control flat, $\sim$78 GPU-h/arm (\S\ref{sec:rlvrresults}) \\ \rowsep
Run a random-reward control arm with any RLVR claim & it reproduced our baseline on every axis; without it, every gain here would be unattributable (\S\ref{sec:rlvrresults}) \\
\midrule
\textbf{NOT supported} & \textbf{why} \\
\midrule
A specific corpus version & five expansions, all inside the noise floor \\ \rowsep
A data-selection method & selection $\approx$ random, $-0.5$ pp ($0.31\sigma$) \\ \rowsep
A LoRA rank, stride, or epoch count & never varied; no evidence either way \\ \rowsep
A per-domain model choice & two of three per-domain score spreads exceed the seed range itself (\S\ref{sec:domain}) \\
\bottomrule
\end{tabular}
\caption{Supported and unsupported configuration choices, each with its evidential basis.}
\label{tab:recommend}
\end{table*}

\section{A third pre-registration: verifiable-reward RL, designed before its numbers}\label{sec:rlvr}

Everything this paper measures ends at the boundary of imitation. SFT reproduces the traces it is
shown, and three of our measured defects are structural to that: the released Qwen checkpoint
fails to emit the requested answer line on $24\%$ of Greek items and leaks the answer into the
trace channel on $3.5\%$ (\S\ref{sec:metrics}), and clean demonstration plausibly cannot push a
leak \emph{rate} to zero, because the imitation objective never sees the counterfactual. The third
is the one \S\ref{sec:override} leaves without a mechanism: after language-matched training the
explicit override instruction is obeyed on $44.8\%$ of items on this family, and we do not know
what makes obedience trainable. All three quantities are deterministically checkable, which makes
them exactly the setting for reinforcement learning with verifiable rewards
(RLVR)~\citep{deepseekai2025r1,shao2024deepseekmath}: the objective can be written without a
learned reward model or an LLM judge, so the reward cannot drift, cannot be sycophantic, and can
be audited line by line. This section pre-registers that experiment the way \S\ref{sec:plan}
pre-registered the paper's two questions: design, reward, defenses and decision rules are frozen
here, and were frozen (rules, reward, fuzz suite and arms) \emph{before} the first optimizer
step. The outcomes, measured after that freeze, close the section (\S\ref{sec:rlvrresults}).

The risk the design must carry is also on record: GRPO-style optimization is a known amplifier of
cross-lingual drift, with reasoning reverting to the model's dominant language as
training progresses~\citep{park2025collapse}; at small dense scale, an accuracy-only reward
collapsed language consistency to zero, and a language-consistency term repaired it at no accuracy
cost~\citep{zhang2025thinknatively}. Whether that repair holds when the property being protected
was installed by SFT, on a mixture-of-experts model, through a LoRA adapter, is untested, and our
headline result (\S\ref{sec:langmatch}) is precisely what an accuracy-only gradient would erode
first.

\textbf{Objective and environment.} GRPO~\citep{shao2024deepseekmath} on the released Qwen
checkpoint: $8$ sampled completions per prompt, group-relative advantage, no KL term
($\beta{=}0$, following~\citealp{yu2025dapo}), constant learning rate $10^{-6}$, temperature
$0.7$, completion budget $1{,}536$ tokens. The budget is a measured decision, not a default: at
$1{,}024$ tokens roughly $90\%$ of completions truncated, and because a truncated completion is
gated to zero correctness (below), the correctness term was starved and the run optimized
termination and format only. Adaptation is LoRA~\citep{hu2022lora} at $r{=}32$, $\alpha{=}64$ on
the fused expert projections, the same fused-parameter targeting the SFT used
(\S\ref{sec:families}), with dropout fixed at $0$: under a policy-gradient objective, dropout
makes the update policy differ stochastically from the rollout policy and corrupts the importance
ratio, so a regularizer that is habit in SFT is a correctness bug here. All reward arms run
simultaneously on identical hardware with the same data order and the same seed, so the only
factor that varies between arms is the reward.

\textbf{The prompt pool.} Problems enter in language-matched \emph{pairs}, the same item once in
Greek and once in English, exactly as the SFT corpus was built (\S\ref{sec:langmatch}), because a
Greek-only pool with any language-shaped term aims a forgetting gradient at the paper's central
property. A held-out slice carries the explicit override instruction, with its phrasing copied
verbatim from the evaluation probe of \S\ref{sec:override}: both contradicting directions
(Greek question instructing English, and the reverse) \emph{and} both agreeing controls, so the
learnable target is the instruction, not the flip. Gold answers are numeric only, verified by the
locale-aware parser below; the pool is deduplicated and decontaminated by $13$-gram overlap
against both the SFT corpus and every evaluation benchmark, in both languages
(\S\ref{sec:lied} is why in both).

\textbf{A reward with no judge.} Five deterministic terms: (i)~\emph{correctness}, the anchored
answer equals gold under a locale-aware numeric parser (in Greek convention $17.500$ is
seventeen and a half thousand and $3{,}5$ is three and a half, and \S\ref{sec:locale} documents a
scorer of ours that misread exactly this); (ii)~\emph{language consistency}, trace matches the
question's language (redirected to the \emph{instructed} language whenever an explicit
instruction is present, so the two terms cannot both be satisfied by ignoring the instruction);
(iii)~\emph{format}, the requested answer line is present; (iv)~\emph{termination}, generation
ends inside the budget (a truncated completion scores zero correctness, because a right number
inside an unfinished trace is not a solved problem); and (v)~\emph{override obedience} on the
instructed slice. Two gates run underneath: language terms require a minimum of letters after
stripping code and \LaTeX{} (\S\ref{sec:locale}); and every behavioural term requires the trace to
contain \emph{work}, an intermediate value that is not the answer restated, deliberately not a
length test.

\textbf{Adversarial pre-flight, because a found exploit is permanent.} A policy that discovers a
reward bug encodes it in weights; unlike an analysis bug, it cannot be fixed afterwards. So the
reward ships with an adversarial suite of $28$ attacks that must pass before any rollout:
trace elision (answer line only), fluent Greek filler with no computation and its
$10\times$-length variant, answer shotguns and number sprays against the extractor, code-block
laundering of English content, the locale exploit above, truncation mid-number, disobeying and
obeying override pairs (the disobeying trace must score strictly lower), and a robustness family
that must not crash. Building this suite caught three exploits in our own draft reward before
training (a fluent no-computation trace at full marks among them) and motivated the work gate.
One failure mode the suite structurally cannot catch is a term whose \emph{input} never arrives:
a reward called without its instruction field silently scores the term at zero and the arm
degenerates into a different arm with no error raised; we caught exactly this in a pilot, and the
trainer now refuses to start if the override arm's pool carries no instructed rows, and asserts
after training that instructed rows actually reached the reward. During training we store every
rollout (prompt, all $8$ completions, and the per-term score breakdown), so any suspicious
number can be traced to the text that earned it.

\textbf{Design: one varied factor, and a control that can veto everything.} Four arms, identical
in data, steps, seed and hardware, differing only in active reward terms: correctness{+}format
{+}termination (the collapse probe: what does an accuracy-shaped gradient cost the language
property?); the same plus language consistency (the candidate recipe); the same plus override
obedience on the instructed slice (the mechanism probe for \S\ref{sec:override}); and a
\emph{random-reward control}: same steps, same data, reward drawn uniformly at random. The
control is not decoration: random rewards have been shown to recover most of an RLVR gain on this
model family~\citep{shao2025spurious}, so any axis on which the control matches a trained arm is
elicitation, not learning, and the pre-registered rule is that the result on that axis is
withdrawn regardless of what the trained arms show.

\textbf{Decision rules, frozen before the numbers.} Fidelity is a \emph{gate}, not a tradable
term: no outcome that buys a metric with Greek-trace fidelity below $97.0\%$ counts as a win.
Format fallback must reach ${\leq}10\%$ and answer-channel leak ${\leq}1.5\%$ to claim that RLVR
closes defects SFT did not ($24\%$ and $3.5\%$ baselines); the override question carries a
three-branch rule: a rise below $+5$\,pp is the null, a rise of ${\geq}15$\,pp reaching
${\geq}60\%$ from $44.8\%$, with the reverse direction held at ${\geq}80\%$, is the only outcome
that may be called \emph{trainable}, and anything between is reported as reward-responsive with
the missed bar stated (the full frozen rule table is archived with the released artifacts);
accuracy movements inside the $\pm7.7$\,pp seed floor of
\S\ref{sec:variance} are reported as inside the floor, never as gains; and every claim is
measured on the held-out instruments of \S\ref{sec:metrics}, never on the training reward, which
is the quantity a reward hack inflates. The null branches are pre-committed too: if the floors do
not drop, the defects are reported as not reward-addressable at this budget; if override does not
move, the finding of \S\ref{sec:override} keeps its ``no mechanism'' framing, strengthened.
Abort conditions (entropy collapse, response length exceeding twice baseline, reward rising with
fallback, any term saturating early) stop a run before it can manufacture a result.

\subsection{The outcomes, against the frozen rules}\label{sec:rlvrresults}

The round ran as designed: $2{,}000$ prompts ($800$ per language, paired, plus a $400$-row
override slice), one epoch, $8$ completions per prompt, four arms in parallel on identical
hardware with the same seed and data order, $\sim$$39$ hours. No abort condition fired; the
fuzz suite passed before launch; the override slice's arrival at the reward function was
asserted at run end on every arm. Table~\ref{tab:rlvr} holds every number a decision rule reads.

\begin{table*}[t]\centering\footnotesize
\setlength{\tabcolsep}{3.4pt}
\begin{tabular}{@{}lrrrrrr@{}}
\toprule
 & Greek & & fall- & & override & reverse \\
arm & fid. \%$\uparrow$ & acc$\uparrow$ & back \%$\downarrow$ & leak \%$\downarrow$ & el$\to$EN \%$\uparrow$ & en$\to$EL \%$\uparrow$ \\
\midrule
SFT checkpoint (before) & 97.98 & 73.7 & 24.1 & 3.53 & 44.8 & 83.7 \\
random reward (control) & 98.06 & 74.0 & 22.1 & 3.61 & 44.1 & 83.8 \\
correct+format+term.    & 98.22 & 77.2 & \textbf{2.5} & \textbf{0.00} & 48.8 & 84.7 \\
\;\;+ language          & 98.02 & 77.2 & 5.6 & 0.04 & 45.9 & 84.5 \\
\;\;+ override          & \textbf{98.27} & 77.0 & 2.8 & 0.02 & \textbf{53.9} & \textbf{85.7} \\
\bottomrule
\end{tabular}
\caption{The pre-registered RLVR round, scored on the held-out instruments (Greek think lane
$n{=}5{,}156$; override lanes $n{=}1{,}000$/$1{,}100$), same scorer, same day. The \emph{before}
row is the released checkpoint's frozen dump rescored with the current scorer; its full same-day
regeneration reproduced all $5{,}156$ responses \emph{bit-identically} (greedy decoding is
deterministic here), and the plain-mode lanes to the decimal ($44.8$/$83.7$): the before column
is not a provenance caveat but a verified constant. The control row is
the entire admissibility argument: on every axis the random-reward arm reproduces the baseline,
so nothing below can be elicitation. Accuracy is shown for completeness only: $+3.3$--$3.5$ is
inside the $\pm7.7$ seed floor of \S\ref{sec:variance} and is claimed by nobody.}
\label{tab:rlvr}
\end{table*}

\textbf{The floors drop, and the control proves it is learning.} Fallback $24.1\%\to2.5\%$ and
answer-channel leak $3.53\%\to0.00\%$ on the plain-RLVR arm, both past their pre-registered
thresholds ($\leq$$10\%$, $\leq$$1.5\%$), with fidelity held. The random-reward control (same
steps, same data) moves on neither ($22.1\%$, $3.61\%$). The branch fires as written in
advance: \emph{in this round, RLVR closed the format and leak defects that our SFT did not},
and for the leak, plausibly could not (the counterfactual argument above), at $\sim$$78$
GPU-hours per arm ($39$ hours wall on two GPUs) over a $2{,}000$-prompt pool. On format, SFT is
not helpless (a targeted repair dose cut another family's fallback $70{\to}26\%$,
\S\ref{sec:plan}), but it plateaued at $24\%$ on this one, where the reward reaches $2.5\%$.

\textbf{Steerability is reward-responsive; the trainability bar was not met.} The override arm
moves instructed compliance $44.8\to53.9\%$ ($+9.1$\,pp; the control sits at $44.1$, so the
control-adjusted effect is $+9.8$\,pp at $n{=}1{,}000$ per cell) while \emph{holding} every
hold-gate: reverse-direction compliance rises to $85.7\%$ (rule: $\geq$$80$), the agreeing
direction is untouched ($98.5\%$, unchanged from baseline; same probe as
Table~\ref{tab:override}, not shown in Table~\ref{tab:rlvr}), fidelity is the highest of any arm
($98.27$). Under the frozen three-branch rule this is the middle branch, and we word it
accordingly: the channel \emph{responds} to a verifiable reward (the first mechanism-bearing
movement of this quantity), but the $\geq$$+15$\,pp\,/\,$60\%$ trainability bar was missed, so
\emph{trainable} is a claim this round does not earn. One ordering we report without a
mechanism: the +language arm gains least ($45.9$ vs $48.8$ without the language term), consistent
with a language-consistency pressure on the bulk of the pool opposing instructed switches, even
though the reward redirects that term on instructed rows by construction.

\textbf{The collapse did not come.} The arm with no language protection ended at $98.22\%$
Greek-trace fidelity, above its starting point, with zero in-question switches. At this
scale and budget (a MoE adapted by LoRA, $16{,}000$ completions, the property installed by SFT
rather than by RL), the accuracy-shaped gradient did not erode language matching. We state this
as bounded: it contradicts the dense-scale prior~\citep{park2025collapse,zhang2025thinknatively}
at our operating point, not everywhere.

\keyfinding{\textbf{Finding 5.} The pre-registered RLVR round moved what SFT did not, and the
control shows it is learning: answer-format fallback $24.1\to2.5\%$ and answer-channel leak
$3.53\to0.00\%$, both past their frozen thresholds with fidelity held, while the random-reward
arm, trained on the same steps and data, reproduced the baseline on every axis. The override
channel responded ($+9.1$\,pp, every hold-gate kept) but missed the frozen trainability bar; the
language property survived an unprotected accuracy gradient.}

\textbf{Disclosures.} The gates lanes were re-run mid-evaluation after the same-day baseline
exposed a render flag silently disabling the reasoning trace: the sixth entry of
\S\ref{sec:lied}, and the reason the \emph{before} column's provenance is stated in the caption.
The override slice lives in the shared pool, so the +language$\to$+override contrast is a dose
comparison on one term, not presence/absence. And the scope is one family (Qwen), one seed per
arm, the final checkpoint, an unscreened pool: these are existence proofs with a control, not
recipes. The override arm passed every hold-gate (fidelity, both held directions) while missing
the $60\%$ override target, and ships as
\textbf{Sophea-Qwen3.6-v1.1}, alongside the SFT release it refines.

\section{What to take away}\label{sec:takeaway}

\textbf{For practitioners.} Run the seed control before the ablations. One extra training run told us
more than five corpus versions. Report \texttt{fallback\%} or an equivalent compliance rate alongside
accuracy; ours separated ``cannot reason'' from ``will not answer in the requested form'' and supplied
the mechanism for our only surviving effect. Keep a human-translated subset of any machine-translated
benchmark.

\textbf{For the low-resource setting specifically.} A fine-tune that leaves accuracy unchanged is not
a failed fine-tune. Ours made a model reason in Greek instead of English, with budget adaptation the
base does not have, and, on the family whose traces shorten enough to repay Greek's token
fertility, at $3\times$ fewer tokens; on the others the trace-language gain costs token parity or
a $1.6\times$ premium (\S\ref{sec:quality}). Those are the properties a deployment cares about, and
an accuracy table cannot see any of them.

\textbf{What we do not claim.} No fine-tuned arm beats its base on the pooled Greek
\emph{reasoning} benchmark in any family (the NLU retention suite is a different quantity, where
two arms gain, Table~\ref{tab:forgetting}), and we
report that as a measured null with a stated noise floor rather than an unresolved comparison. We
report the fine-tunes' behavioural advantages, and we report that the corpus work which consumed most
of the project produced no measurable accuracy effect at all.

\section{Related work}\label{sec:related}

\textbf{Reasoning language.} Which language a multilingual model \emph{should reason in} has an
established literature, and the default answer is English. \citet{shi2023mgsm} show that
English-language chain-of-thought outperforms native-language chain-of-thought on multilingual
math, and \citet{etxaniz2024better} report the same direction under self-translation: translate to
English, reason, translate back. \citet{wendler2024llamas} give it a mechanism: multilingual
transformers' latent space works in an English-like intermediate representation. Prompting
interventions ride this current rather than oppose it: cross-lingual-thought
prompting~\citep{huang2023xlt} machine-translates the problem into English and reasons there, and
the translation-distillation line extends English CoT supervision to non-English tasks at
scale~\citep{chen2024mathoctopus,qin2023xrc}. The closest point on the map is
DeepSeek-R1~\citep{deepseekai2025r1}, whose RL phase adds a language-consistency reward precisely
because unconstrained RL drifted into language mixing; the reward keeps the trace in the prompt's
language at a small measured cost to reasoning performance, treating a prompt-language trace as
the success case rather than the failure, as does the consistency-reward work that
followed~\citep{zhang2025thinknatively,park2025collapse}. The prompting and
distillation work above makes the opposite call: reasoning in English is the success case and
drift back to the prompt's language the failure. Our deployment premise sides with R1's trade and
goes further: for a low-resource-language deployment of a reasoning model (here Greek) the trace
\emph{is} the product, and
an opaque English trace is the defect, not the target. What we add over R1's reward is the
setting and the measurement: the property is installed by SFT alone in a mid-resource language,
and we measure both directions of the resulting steering asymmetry (the locked fine-tune obeys
``reason in English'' on $0\%$ of items; the base ignores ``reason in Greek'' on $72\%$,
\S\ref{sec:langmatch}); that
combination is where this paper sits.

\textbf{Seed variance and reporting.} The call to report variance rather than a best run is at
least as old as the benchmark culture it critiques: \citet{dodge2020show} argue expected-validation-
performance budgeting, \citet{bouthillier2021accounting} show baseline reorderings disappear once
all sources of training randomness are marginalised, and \citet{madaan2024variance} quantify how
evaluation choices (prompt, seed, few-shot draw) reorder LLM leaderboards. Our contribution to that
line is not the call but two measured transfers: the effect size at \emph{MoE + LoRA + a
mid-resource language} scale (7.7 pp on the identical configuration, sd 4.4 pp, \S\ref{sec:variance}),
and the asymmetry that the seed variance lands almost entirely on accuracy while our behavioural
metrics sit flat across the same three seeds. The second transfer is what lets a paper this size
survive its own noise floor: it is also the thing we have not seen measured in the seed-variance
literature, which works almost entirely in accuracy or expert-human preference space.

\textbf{Trace structure, selection, and the self-taught lineage.} Our corpus-generation pipeline is
a direct descendant of STaR~\citep{zelikman2022star}: sample a trace, keep it only if the final
answer agrees with gold, discard the rest rather than repair them. What the trace itself should
\emph{look like} is laid out by \citet{li2025structure} (``the structure of Long CoT is critical
to the learning process, whereas the content of individual reasoning steps has minimal impact'')
and \citet{gandhi2025cognitive}, whose reasoning-behaviour count (verification, backtracking,
subgoal setting, backward chaining) names the variable our structure score was groping for. The
cost side of traces has its own literature: the over-thinking phenomenon in long-CoT models
\citep{chen2025overthink} is what our budget-overrun metric (\S\ref{sec:quality}) quantifies per
item rather than per benchmark. Our data-selection null is anticipated by \citet{xia2024random},
who found that self-scoring selection methods ``struggled to significantly outperform random
selection'' at scale: the control we should have run first, and did run last.

\textbf{Tokenizer fertility and the word-vs-token trap.} That low-resource languages pay a
tokenizer tax is a standard observation in multilingual evaluation; \citet{lu2023fertility}
quantify fertility gaps of the order we measure ($2.3$--$2.5\times$ for Greek against English on all
three families, \S\ref{sec:quality}). The figure matters here because it is the hidden hand behind
every cost comparison this paper makes: a word-denominated reasoning budget is the language-fair
metric, and a token-denominated one silently double-counts the fertility tax (M3,
\S\ref{sec:metrics}; we report both). The same trap, units quietly re-priced per language,
recurs wherever a derived
quantity (cost per correct answer, budget overrun) inherits the units of a primary one.

\textbf{The Greek line, and our instruments' provenance.} Greek NLP has moved past the era of
translating nothing and measuring nothing: Meltemi~\citep{voukoutis2024meltemi} and
Krikri~\citep{roussis2025krikri} are the dedicated open Greek base models, and their evaluation
suites (Greek translations of MMLU, ARC, HellaSwag, WinoGrande and
Belebele~\citep{bandarkar2024belebele}) are the same instruments our retention suite
(\S\ref{sec:langmatch}) inherits. Sophea-Titan-1, the model whose instruction corpus is
our direct half, belongs to that same line, as do our two companion efforts: a Greek adaptation of
the Nemotron retrieval stack with HERA, a large-scale Greek retrieval-augmented-generation
benchmark~\citep{kirouane2026nemotron}, and MORFES, an expert-verified benchmark for productive
Greek inflectional morphology~\citep{perros2026morfes}. The threats to the axes we evaluate \emph{on} are
equally documented, and two of ours (\S\ref{sec:lied}) are new exemplars of known mechanisms:
\citet{artetxe2020translation} show that translating premise and hypothesis independently reduces
lexical overlap in NLI (our degrading category), and \citet{singh2024globalmmlu} show that
rankings change depending on whether models are evaluated on the full or the culturally-sensitive
subset of a translated benchmark. What \S\ref{sec:lied} adds to that literature is a distinct
mechanism: the artifact can favour the \emph{fine-tuned} model rather than distorting all models
alike, when the benchmark translator and the training-data generator are the same model family.
Degenerate repetition under greedy decoding is the classic result of
\citet{holtzman2020curious}; our contribution is negative, that it is not separable from length in
this setting.

\section*{Limitations}\label{sec:limits}

\textbf{Every model here is a sparse MoE, and every one is adapted with LoRA.} Every result here is therefore
a statement about \emph{LoRA on mixture-of-experts models}, not about supervised fine-tuning in
general. Two specific reasons this could matter rather than being boilerplate: the adapters touch
only a strided subset of MoE layers plus the shared expert, so a dense model (where LoRA reaches
every FFN) may behave differently; and the seed sensitivity we report (\S\ref{sec:variance}) has a
plausible MoE-specific mechanism, since routing decides which experts receive gradient at all, and a
different initialisation can send a different subset of experts down a different path. We did not
test a dense baseline, and cannot separate ``LoRA SFT is seed-sensitive at this scale'' from
``sparse routing amplifies seed sensitivity''. A dense control is the single cheapest experiment
that would sharpen this paper.

Single seed for all but one configuration; the noise floor is estimated from $n{=}3$ on one arm and
$n{=}2$ on another. Four conditions were never varied: LoRA rank and targets, one epoch, the fixed
non-reasoning arm in every two-phase mix, and greedy decoding. All three axes are Greek versions of
English benchmarks; a natively-authored Greek commonsense probe we built came out at ceiling
($96$--$97\%$ for every arm) and could not discriminate. The logic axis is the smallest after
decontamination and carries the seed instability, and every arm is near-blind to one of its three
classes ($14$--$16\%$ recall on \emph{False}), a failure a single accuracy number conceals.

\textbf{Greek is mid-resource, and the auditability claim is a hypothesis.} Greek is an EU
official language with dedicated open models; whether this recipe transfers to truly low-resource
languages (which lack the strong base and the language-directed corpus it depends on) is
untested. And no human read the traces: apart from the $\sim$$150$-trace hand-labelled switching
probe of \S\ref{sec:setup} itself (one annotator, no guidelines or agreement measurement),
every fidelity number in this paper is automatic, so
``a trace the user can read and audit'' is measured here only as script identity (the small
LLM-judged register and grammar probes of \S\ref{sec:forget} are the sole exception), not as fluency,
terminology quality, or followability; and it inherits the standing assumption that a
chain-of-thought trace is faithful to the computation it narrates. Two cheaper alternatives to SFT
were not compared: translating the base's English trace post hoc, and few-shot Greek-trace
exemplars (our ``prompting cannot reach it'' evidence covers bare instructions only). Finally, the
zero-switch target treats any code-switching as a defect; Greek technical registers routinely
borrow English terms, and we did not measure what the monolingual constraint costs in terminology
fidelity.

\subsection*{Availability}

We release five fine-tuned checkpoints across the three families of Table~\ref{tab:models}:
\textbf{Sophea-Qwen3.6-v1}, \textbf{Sophea-OSS-v1}, \textbf{Sophea-Nemo-3-Nano-v1} and
\textbf{Sophea-Nemo-3.5-Lightning-v1} (the four SFT releases), plus
\textbf{Sophea-Qwen3.6-v1.1}, the RLVR override arm of \S\ref{sec:rlvrresults}, which passed
every hold-gate (fidelity $98.27$, both held directions kept; the $60\%$ override target itself
was missed). The Qwen releases ship the base's full multimodal (vision) stack and its
multi-token-prediction head for speculative decoding; the Lightning release ships its base's
MTP head as well. All five ship in the \textbf{Sophea Reasoning Models} collection at
\url{https://huggingface.co/collections/KIEFERSA/sophea-reasoning-models}.

The controls are the methodology we would most like reused (\S\ref{sec:takeaway} and
Table~\ref{tab:recommend} carry the full list). Each is a few hours of compute, none needs
our models or our corpus, and each changed a conclusion we had already written down; run in the
order given they cost less than one ablation and would have saved us most of a project.

\balance
{\small
\bibliographystyle{plainnat}
\bibliography{refs}
}

\end{document}